\documentclass[nonacm]{jair}
\usepackage[utf8]{inputenc}

\setcopyright{cc}
\copyrightyear{2026}
\acmDOI{10.1613/jair.1.x}

\RequirePackage[
  datamodel=acmdatamodel,
  style=acmauthoryear,
  backend=biber,
  giveninits=true,
  uniquename=init
  ]{biblatex}

\newcommand{\citeay}[1]{\citeauthor{#1} \citeyear{#1}}

\usepackage{dialogue}
\usepackage{calc} 
\usepackage{comment}
\usepackage{siunitx}

\newlength{\widestname}
\makeatletter
\renewenvironment{dialogue} {%
    \begin{list}{} {%
        \setlength\itemsep{\z@ \@plus .5ex} %
        \setlength{\parsep}{\parskip} %
        \setlength{\rightmargin}{0pt} 
        \setlength{\labelwidth}{\widestname} 
        \setlength{\labelsep}{0.5em} 
        \setlength{\leftmargin}{\labelwidth} 
        \addtolength{\leftmargin}{\labelsep} 
        \defcommand\speak [1] {\item[{##1}]} 
        
      }%
      \PreDialogue\relax
    }{%
  \end{list}%
  }
\makeatother

\begin{document}


\title{Robotics-Inspired Guardrails for Foundation Models in Socially Sensitive Domains}
\author{Rebecca Ramnauth}
\authornote{Corresponding Author.}
\orcid{0000-0003-3556-532X}
\email{rebecca.ramnauth@yale.edu}
\affiliation{%
  \institution{Yale University}
  \city{New Haven}
  \state{Connecticut}
  \country{USA}
}

\author{Dra\v{z}en Br\v{s}\v{c}i\'{c}}
\orcid{0000-0001-8477-6460}
\email{drazen@i.kyoto-u.ac.jp}
\affiliation{%
  \institution{Kyoto University}
  \city{Kyoto}
  \country{Japan}}

\author{Brian Scassellati}
\orcid{0000-0002-7671-7759}
\email{brian.scassellati@yale.edu}
\affiliation{%
  \institution{Yale University}
  \city{New Haven}
  \state{Connecticut}
  \country{USA}
}

\renewcommand{\shortauthors}{Ramnauth et al.}

\begin{abstract}
Foundation models are increasingly deployed in socially sensitive domains such as education, mental health, and caregiving, where failures are often cumulative and context-dependent. Existing guardrail approaches---ranging from training-time alignment to prompting, decoding constraints, and post-hoc moderation---primarily provide empirical risk reduction rather than enforceable behavioral guarantees, and largely treat safety as a property of individual outputs rather than interaction trajectories. We reframe guardrails as a problem of runtime behavioral control over interaction trajectories, drawing on robotics to introduce formal constructs for constraint enforcement in uncertain, closed-loop systems. We instantiate these ideas in the Grounded Observer framework and apply it across three real-world deployments: small talk, in-home autism therapy, and behavioral de-escalation in schools. Across settings, the framework enables runtime interventions that mitigate drift into undesirable interaction regimes while adapting to diverse social contexts. We discuss extensions to the framework and propose research directions toward stronger guarantees.

\end{abstract}

\maketitle

\section{Introduction}

Foundation models are rapidly being integrated into various fields, from medical diagnostics and financial predictions to socially sensitive areas such as education, mental healthcare, and support for individuals with disabilities. Despite being aware of the inherent risks of AI hallucinations, misinformation, and bias, a recent large-scale global study revealed that 66\% of respondents are still willing to use this nascent technology in sensitive areas such as personal advice and relationship counseling ~\cite{capgemini_ai}. This tension highlights both the promise and the fragility of deploying foundation models in high-stakes settings: demand for these systems is growing faster than our ability to bound their behavior. Simultaneously, widespread adoption remains constrained by a central technical gap, yet we lack guardrails that can reliably prevent undesired behavior and enforce stable performance under real-world uncertainty. 



In fields where accuracy and reliability are paramount, such as healthcare and finance, the consequences of errors can be severe. Conversely, in socially sensitive domains, where the criteria for success are less formalized and outcomes are mediated through emotion and interpersonal norms, missteps can be as consequential. For example, a system intended to provide calming techniques in a clinic waiting room could inadvertently intensify distress if it delivers generic or socially miscalibrated guidance. A response that fails to recognize urgency or emotional context may be perceived as dismissive, undermining trust and worsening anxiety rather than alleviating it. These risks motivate the need for robust guardrails that can reliably constrain behavior in real time, protecting both users and the integrity of the interaction.


Although guardrails for foundation models have rapidly become a core concern in machine learning, the technical conversation around them remains fragmented. Much of the literature treats safety as a property that can be embedded into a model through training: by curating datasets, applying reinforcement learning from human feedback, tuning system prompts, or filtering outputs with classifiers ~\cite{dong2024building, dong2025safeguarding, shen2024towards, ji2023ai}. These methods have led to important practical gains, but they also reflect an implicit assumption---that behavioral reliability is primarily a matter of shifting a model's output distribution toward desirable responses. This assumption becomes increasingly fragile as foundation models are deployed not as static question-answering tools, but as autonomous components embedded in interactive systems that operate continuously, adapt to user behavior, and influence real-world decisions ~\cite{shen2024towards}. In such settings, the relevant failure modes are not always catastrophic single-turn violations ~\cite{harris2016emergent, slota2020good}. Instead, harmful behavior often emerges gradually: a system drifts toward inappropriate conversational roles, accumulates subtle norm violations, escalates affect, or produces interactions that are locally acceptable yet globally destabilizing. If foundation models are to be deployed in socially sensitive domains such as education, healthcare, therapy, and caregiving, then the question is no longer merely whether a model can generate safe responses on average, but whether an interactive system can reliably remain within behavioral bounds over time.

Designing usable systems that impose limits on foundation models involves two key challenges. First, foundation models are learned statistical policies. Their behavior emerges from high-dimensional pattern induction over vast training corpora rather than from explicit symbolic rules. As a result, it is difficult to express human constraints directly within the model's internal decision process. In contrast, traditional rule-based systems rely on symbolic representations that are compositional and amendable to formal reasoning, but are not naturally aligned with the latent representations that govern foundation model outputs. Bridging this gap requires translating high-level human concepts (e.g., ``be empathetic,'' ``do not escalate conflict,'' or ``avoid medical advice'') into operational criteria that can be evaluated reliably at runtime. This translation problem reflects a mismatch between the statistical nature of learned models and the symbolic structure of the limits we wish to enforce. While neurosymbolic approaches attempt to reconcile these paradigms by combining learning-based representations with symbolic constraints or reasoning modules (e.g., \citeay{garcez2023neurosymbolic}), robust and scalable integration remains an open problem, particularly in socially sensitive settings where constraints are context-dependent and often only partially observable. 

Second, even if constraints can be specified, they must be enforced under dynamic, user-specific conditions. Foundation models are increasingly deployed in interactive settings where the appropriate behavior cannot be captured by static rules alone. The same response may be supportive in one context and harmful in another. A static rule set may prohibit obvious violations (e.g., self-harm encouragement), yet still fail to regulate more subtle trajectory-level failure modes, such as gradual drift into inappropriate advice-giving, overconfidence, or emotionally mismatched reassurance. Consequently, guardrails for foundation models must support real-time behavioral adaptation, enabling systems to adjust their action selection policies in response to evolving interaction state and individualized user needs \cite{wang2023adapting,chen2024large}. This requirement shifts the problem from specifying a fixed set of rules to designing mechanisms that can enforce context-sensitive constraints continuously over long-horizon interactions \cite{raman2022planning}.


These two challenges are not unique to foundation models. Robotics offers a mature technical tradition for addressing precisely this class of problem: controlling autonomous systems under uncertainty while enforcing constraints over trajectories. In action selection for robot systems, an agent must decide on actions to take, often using large-scale statistical models, while adhering to user-specified rules, such as ``don't touch the stove.'' In robotics, the learned or optimized policy may produce effective behavior on average, yet still require runtime enforcement to prevent unsafe actions in rare or adversarial states. This has motivated a family of techniques that introduce explicit supervisory layers into the control loop. For example, shielding techniques prevent particular actions from being executed ~\cite{alshiekh2018safe}, effectively restricting the robot's behavior, while interactive policy shaping modifies the action selection policy in real time based on user input or situational changes ~\cite{griffith2013policy}. These approaches provide a concrete template for reconciling the flexibility of statistical decision-making with the rigidity of constraint satisfaction: rather than attempting to encode all constraints into the policy itself, they impose an external locus of control that can filter or override actions at runtime ~\cite{biza2021action} 


 Furthermore, in robotics, safety is rarely treated as a property that can be learned once and assumed thereafter. Instead, autonomous systems are typically structured around explicit models of state and action dynamics, and are paired with runtime mechanisms that monitor behavior, enforce invariants, and intervene when constraints are violated. The robotics community has long recognized that even high-performing controllers require external safety layers, because the environment is unpredictable, sensing is imperfect, and small deviations can compound into unacceptable outcomes. While foundation models differ from physical robots in important ways, the underlying technical problem is surprisingly similar: both are high-dimensional policies operating in partially observed environments, and both require stable constraint enforcement over time rather than one-shot correctness. Yet despite this conceptual alignment, guardrail research for foundation models has largely evolved in isolation from robotics, often rediscovering ideas that have been formalized for decades in autonomous systems.

 This paper argues that building strong guardrails for foundation models requires importing not only intuitions from robotics, but technical constructs that provide an explicit locus of runtime control. We propose that deployed foundation model systems should be treated as constrained dynamical systems: systems whose behavior unfolds over state trajectories and whose safety properties are defined by admissible regions of state-action space. This framing makes explicit what is often implicit in alignment discourse. Specifically, it distinguishes the base policy from the safety mechanism, separates the specification of constraints from the process of satisfying them, and clarifies the assumptions under which any behavioral guarantee can be claimed. It also exposes a key trade-off: guardrails can offer broad coverage but weak enforcement, or strong enforcement but limited scope, and these properties depend not only on the model but on the architecture in which it is embedded.

 To make these ideas concrete, we present the Grounded Observer framework as a practical instantiation of this constrained dynamical systems view for foundation models systems in socially sensitive interactive domains. 
 The framework decomposes a deployed system into a base model that generates candidate actions and an observer model that evaluates these actions against externally specified constraints, enforcing admissibility at runtime through filtering, regeneration, or redirection. Constraints are represented as modular overlays that can be composed and adjusted without retraining the underlying base model, and enforcement is performed over interaction trajectories rather than isolated turns. While the framework does not provide absolute formal guarantees (due to uncertainty in state estimation and the difficulty of operationalizing social norms), it introduces an architectural structure that enables enforceable behavioral bounds to be approximated and audited systematically.

 We demonstrate how this framework operates through three deployment case studies: developing agents that can successfully sustain small talk, enabling in-home autism therapy, and supporting behavioral de-escalation in public schools. These applications were selected not because they represent canonical benchmark tasks, but because they highlight the types of guardrail problems that are increasingly central to real-world deployment: behaviors that are judged over long horizons, shaped by social norms rather than ground truth labels, and sensitive to subtle forms of drift that cannot be corrected through static prompting alone. Across these deployments, the observer architecture was configured in different ways (ranging from soft constraint stabilization in small talk to hierarchical activity-level supervision in de-escalation), illustrating how robotics-derived constructs naturally emerge when foundation models are deployed as autonomous interactive systems.

This paper advances the central claim that an important phase of guardrail research for social AI must move beyond training-time alignment and post-hoc evaluation toward runtime behavioral control. By drawing on robotics, we can develop a formal vocabulary and a set of technical constructs for doing so, along with a concrete architecture that demonstrates how these constructs can be realized in deployed systems. As foundation models become embedded in social environments where failure modes are cumulative, context-dependent, evaluated socially or relationally, and difficult to reverse, developing enforceable runtime guardrails is more than an optimization problem---it is a necessary prerequisite for safe and responsible deployment.

\section{Existing Approaches}

Efforts to build behavioral guardrails for foundation models have evolved across multiple layers of the machine-learning stack and typically reflect a trade-off between \textit{breadth} of coverage and \textit{strength} of guarantees. In this section, we summarize these efforts and organize them by the technical locus of control.

\subsection{Training-Time Alignment} \label{sec:background-training}
The earliest and most widely adopted approaches operate at training time, with the aim of aligning model behavior with behavioral norms and expectations before deployment. Reinforcement Learning from Human Feedback (RLHF) \cite{christiano2017deep} and its variants---most prominently those introduced in InstructGPT \cite{ouyang2022training} and refined in OpenAI's ``helpful-harmless'' alignment pipelines---optimize models to reproduce preferred responses while penalizing undesirable ones. More recent work, such as Constitutional AI (CAI) \cite{bai2022constitutional}, replaces or supplements traditional human feedback with principle-guided self-critique. Specifically, CAI uses a fixed, human-written ``constitution'' of guiding principles that the model consults to critique and revise its own outputs, and then uses those critiques to train preference modes and fine-tune behavior. 
 
 These methods have produced substantial empirical gains in perceived helpfulness and refusal behavior \cite{ouyang2022training,bai2022constitutional}. However, the guarantees they provide remain fundamentally statistical rather than formal \cite{amodei2016concrete}. Alignment datasets sparsely sample an effectively unbounded interaction space \cite{amodei2016concrete}, while human preference judgments encode implicit normative assumptions and annotator biases. \cite{bender2021dangers, gehman2020realtoxicityprompts}. Moreover, RL-based fine-tuning can induce behavioral drift away from the pretrained distribution unless constrained through explicit regularization mechanisms (e.g., KL penalties or trust-region objectives; \citeay{ouyang2022training}, \citeay{schulman2017proximal}). In addition, what counts as ``safe'' is inherently value-laden, and different stakeholders may disagree on what constitutes acceptable or harmful behavior \cite{gabriel2020artificial}. As a result, current alignment pipelines often trade reduced risk for excessive conservatism, and over-refusal of benign or contextually legitimate requests remains a persistent failure mode \cite{bai2022constitutional}.

\subsection{Prompt- or Decoding-Time Alignment}
Beyond training, a substantial body of work discusses decoding-time control mechanisms that enforce constraints dynamically, without requiring model retraining. Gradient-based steering methods such as the Plug-and-Play Language Model (PPLM) perturb hidden activations to increase the likelihood of desired attributes and suppress undesirable continuations \cite{dathathri2019plug}. Related approaches such as DExperts combine an expert model with an ``anti-expert'' model in a product-of-experts formulation to steer generations away from toxic content \cite{liu2021dexperts}. In parallel, work on constrained and grammar-based decoding restricts token sampling to sequences that satisfy lexical or structural constraints, thereby preventing many invalid or unparseable completions \cite{hokamp2017lexically, post2018fast}. Recent libraries such as Outlines operationalize these ideas by compiling formal constraints (e.g., regular expressions or JSON-schema-like structures) into efficient decoding procedures that guarantee syntactic validity of generated outputs \cite{willard2023efficient}.

These methods provide lightweight, controllable guardrails suitable for downstream integration, but they primarily optimize surface-level properties (e.g., syntax, attribute scores) rather than deeper semantic correctness \cite{dathathri2019plug, bender2021dangers}. Moreover, steering can subtly distort meaning or style, which grammar constraints may reduce fluency or faithfulness when the specified grammar is an imperfect approximation of the target domain \cite{hokamp2017lexically, post2018fast, willard2023efficient}.

One of the most common approaches for constraining model behavior is simply to craft an good prompt. Prompting can be viewed as a form of pre-decoding control, since it shapes the model's internal context and probability distribution \textit{before} any tokens are generated. In this sense, prompt engineering can bias the model toward desirable trajectories, but it does not actively intervene during generation. Although prompt-based control has shown empirical success across many application domains \cite{giray2023prompt, mesko2023prompt, white2023prompt}, it provides limited leverage for robust behavioral constraint. These limitations are especially salient in dynamic, socially sensitive settings, and can be summarized as follows.

First, prompt-based control often lacks robustness. Prompts that perform well in a narrow or carefully curated scenario frequently fail to generalize across small contextual shifts, yielding inconsistent or unintended behavior \cite{zhou2022large,huang2024selective}. Accordingly, prompting is highly context-sensitive, such that minor changes in phrasing or ordering can produce qualitatively different responses, making outputs difficult to predict or reliably control \cite{denny2023conversing, dong2024building}. Moreover, prompt engineering is fundamentally limited in its ability to enforce hard constraints. While prompts can request particular behaviors, they cannot guarantee compliance, nor can they ensure strict adherence to safety policies when adversarial or ambiguous inputs arise \cite{niknazar2024building}.

Prompting also does not translate cleanly to real-world behavioral demands, where social interaction is shaped by nuance and shifting interpersonal cues that cannot be fully captured in static textual instructions \cite{leite2013social}. For example, appropriate mental health dialogue requires sensitivity to subtle emotional signals and evolving conversational context---conditions under which prompt-only control can be brittle. Furthermore, prompt engineering does not inherently support temporal constraints, where the desired behavior depends on the sequence and timing of interactions ~\cite{lyu2024keeping,chen2023forgetful}. For instance, a model prompted to act as a tutor may initially scaffold explanations appropriately, but drift toward overly advanced or repetitive responses over successive turns. Likewise, a mental-health support agent may be instructed to avoid escalating language and to recommend professional resources when self-harm risk emerges, yet fail to apply this policy consistently when risk signals accumulate gradually across several turns. In such cases, shaping initial context alone is insufficient for ensuring stable, long-horizon control.

\subsection{External Safeguards} \label{sec:background-external}
Complementary to the aforementioned internal controls are external moderation models that act as wrapper-level filters around the foundation model, screening either user inputs, model outputs, or both \cite{hurst2024gpt}. OpenAI's Moderation API \cite{openai2022moderation} and Meta's Llama Guard \cite{inan2023llama} are representative examples: they classify input prompts and model outputs into predefined safety categories (e.g., self-harm, hate speech, sexual content) and block or flag violations in real time. These classifiers are easily updated and modular, making them attractive for large-scale deployment or rapid iteration. However, their effectiveness is fundamentally bounded by discrete taxonomies and finite labeled datasets \cite{openai2022moderation, inan2023llama}. Consequently, their coverage can be incomplete, especially for nuanced harm, multilingual content, or implicit bias, and their calibration thresholds often trade recall for precision, leading to false negatives or overly conservative refusals \cite{gehman2020realtoxicityprompts, vidgen2021learning}. 

\subsection{Application-Level Frameworks}
At the application level, guardrails increasingly take the form of programmable safety frameworks that combine multiple validators and safety checks. NVIDIA's NeMo Guardrails \cite{nemoguardrails, rebedea2023nemo} introduces Colang, a domain-specific language for specifying conversational flows with input/output ``rails,'' topic routing, and customizable validators (e.g., hallucination detectors or filters for personally identifiable information). The Guardrails-AI library \cite{guardrailsai} offers a schema-based approach that enforces type and factuality constraints through declarative validators and streaming repair functions. These frameworks embody the practical engineering view of safety as composition: rather than a monolithic model guarantee, safety arises from the interaction of multiple, inspectable control points \cite{amodei2016concrete, hurst2024gpt}. 

However, while these application-layer frameworks provide practical compositional safety through modular validators, they typically operate over local input/output constraints rather than enforcing invariants over long-horizon trajectories---precisely the setting where robotics safety frameworks (e.g., shielding and supervisory control) offer stronger conceptual tools, as later discussed in Section \ref{sec:constructs}. Further, the strength of the resulting guarantees depends critically on the reliability of each individual validator. Many of these validators are themselves learned and error-prone, and heavy layering can add latency or induce undesirable user friction \cite{amodei2016concrete, bai2022constitutional}. 

\subsection{Agent/Action-Level Safeguards}
As foundation models evolve into agentic systems capable of invoking external tools or performing real-world actions, a new line of research focuses on action-level control and runtime verification \cite{firoozi2025foundation, ravichandran2025safety}. In this setting, the model is treated as a decision-making policy that selects actions (e.g., tool calls, API requests, database writes, robot commands) that transition the environment from one state to the next. Consequently, safety is no longer defined solely by the content of a generated response, but by whether the agent's executed action sequence satisfies explicit constraints over permissible behaviors. For example, GuardAgent compiles high-level safety specifications into guard code that monitors an agent's planned actions, rejecting or modifying unsafe ones before execution \cite{xiang2024guardagent}. Similarly, VeriGuard formalizes this process through a two-stage verification pipeline: first, offline specification of allowable policies, followed by a runtime monitor that checks each action against that verified policy \cite{miculicich2025veriguard}. These systems offer stronger, more interpretable guarantees than surface moderation, but they require precise, manually engineered specification and incur computation overhead during planning and actuation---challenges that grow with the open-endedness of modern agent frameworks \cite{hu2023toward}. 


\subsection{Robustness and Adversarial Testing}
A related body of work focuses not on designing new guardrails, but on evaluating whether existing safeguards remain reliable. In practice, robustness failures arise along at least three dimensions: (i) factual reliability in knowledge-grounded generation, (ii) adversarial circumvention of safety policies, and (iii) unintended representational harms induced by detoxification and filtering.

First, substantial attention has been devoted to hallucination and factual risk detection, particularly in retrieval-augmented generation (RAG) systems \cite{lewis2020retrieval}. Methods such as SelfCheckGPT \cite{manakul2023selfcheckgpt} estimate hallucination likelihood using self-consistency sampling, while verification models such as Patronus AI's Lynx \cite{ravi2024lynx} evaluate whether a response is supported by retrieved evidence. Tooling frameworks, including NeMo Guardrails \cite{nemoguardrails}, increasingly integrate such verifiers with provenance tracing and structured validation to support auditable, evidence-grounded outputs. However, these mechanisms cannot provide absolute correctness. Self-consistency is not equivalent to truth, and retrieval pipelines introduce additional attack surfaces through retrieval errors, prompt injection, or adversarially constructed evidence chains \cite{inkawhich2023adversarial, chen2025meta}.

Second, a parallel line of research investigates adversarial robustness of safety-aligned models under deliberate jailbreak attempts \cite{yi2024jailbreak}. Attacks such as the Universal and Transferable Adversarial Suffix demonstrate that models can often be coerced into policy violations via systematic prompt perturbations that generalize across prompts and even across model families \cite{zou2023universal}. In response, benchmark suites such as HarmBench \cite{mazeika2024harmbenchstandardizedevaluationframework}, JailbreakBench \cite{chao2024jailbreakbench}, and HELM \cite{liang2022holistic} provide standardized adversarial evaluations and leaderboards for comparing guardrail efficacy across safety categories. The result of these efforts shows even strong defenses exhibit non-trivial jailbreak success rates, and aggressive defensive tuning often trades reduced risk for degraded helpfulness and task performance. Moreover, evaluation coverage continues to lag behind emerging threat models, including tool-use exploits, prompt-injection attacks on external retrieval systems, and multimodal jailbreak strategies \cite{wu2024vulnerabilities, li2023backdoor}.

Additionally, work on toxicity filtering and bias ``detoxification'' shows that guardrails can themselves introduce failure modes when safety is defined through narrow proxy objectives. Datasets such as RealToxicityPrompts \cite{gehman2020realtoxicityprompts} demonstrate that detoxification methods trained on classifier feedback can overgeneralize, suppressing benign language and triggering false positives on contextually appropriate expressions. More broadly, these findings illustrate a recurring challenge for guardrail design which is, when safety categories are specified too coarsely or optimized via imperfect classifiers, systems may trade reduced risk for overly conservative behavior. This trade-off often results in overblocking and reduced utility of the system even when no genuine harm is present \cite{schneider2024foundation, li2024culturellm, tao2024cultural}.

\subsection{Additional Open Challenges}
Across these layers, industrial practice has converged on a layered defense-in-depth architecture: input moderation, intent routing to risk-appropriate modes, structured decoding, output monitoring and self-checking, followed by agent-level validation and continuous red-teaming. While this stack provides pragmatic resilience and modular maintainability, its assurances are primarily empirical. Guardrails detect and mitigate, but rarely guarantee safety; most operate heuristically, lack formal coverage proofs, and require ongoing adversarial evaluation to remain effective. These gaps (between empirical mitigation and enforceable behavioral bounds) define the current frontier of safety engineering for foundation models. Although many deployed systems already incorporate elements such as state tracking, high-level policies, and output monitoring, these mechanisms are rarely unified into a formal structure that supports principled constraint specification and runtime enforcement over trajectories. This motivates our search for frameworks that can adapt mature assurance concepts from adjacent fields such as robotics, where supervisory control and constraint-based safety have long been studied. 


\section{Why Look to Robotics?} \label{sec:robotics}
Despite this rapid proliferation of guardrail mechanisms for foundation models, a common theme emerges across existing approaches: safety is largely treated as an \textit{empirical} property rather than a \textit{guaranteed} one. For example, training-time alignment improves average behavior but offers no protection against distributional shift or adversarial prompting. Decoding-time steering and post-hoc moderation reduce the frequency of undesirable outputs but remain brittle and reactive. Even agent-level safeguards rely on hand-engineered specifications whose coverage is difficult to assess. What is typically absent from most approaches is a principled account of what behaviors are provably prevented, under what assumptions, and for how long.\footnote{Recent work does aim for stronger guarantees, but only under explicit assumptions (e.g., fixed threat models, stationary prompt distributions, restricted action spaces, or verified code monitors; \citeay{miculicich2025veriguard}).} This gap mirrors a stage robotics passed through decades earlier, when increasingly capable autonomous systems outpaced the tools available to reason formally about their behavior.

Robotics offers a comparatively mature body of work precisely focused on this problem: how to endow autonomous systems with behavioral guarantees under uncertainty. Robots, like foundation models, operate in open environments, interact with humans, and are subject to unexpected inputs and perturbations. Crucially, however, robotics has long been forced to control the consequences of failure (e.g., physical harm, equipment damage, or loss of life; \citeay{heinzmann2003quantitative}, \citeay{dawson2023safe}, \citeay{guiochet2017safety}), driving the development of mathematically grounded techniques for safety and stable constraint enforcement \cite{ames2019control, wabersich2021predictive, desai2019soter, althoff2010reachability, siegel2003sense, brunke2022safe}. As a result, robotics does not rely solely on empirical validation or dataset coverage; instead, it has developed formal tools that specify sets of allowable behaviors, mechanisms to enforce them at runtime, and proof techniques to show that violations are impossible under stated assumptions. 

A key distinction between the robotics and foundation-model safety literature lies in how ``autonomy'' is conceptualized. In much of the foundation model discourse, safety is primarily framed as a property of the model itself---learned during training or corrected through post-processing. Robotics, by contrast, has traditionally treated autonomy as a \textit{closed-loop dynamical process}: perception informs decision-making, decision-making produces actions, actions change the environment, and the resulting state produces new observations \cite{bertsekas2012dynamic, khalil1986new, gautier1995identification}. Safety is therefore not a property of a single component, but a property of an evolving interaction between policy and its environment over time. This perspective naturally leads to behavioral architectures that are layered, runtime-aware, and designed to intervene before unsafe states are reached rather than after violations occur (e.g., \citeay{kurup2012can}, \citeay{kotseruba2016review}, \citeay{nguyen2021robust}). 

This closed-loop framing has motivated robotics to developed a precise language for reasoning about constraints. Concepts such as safe sets, reachability, barrier functions, and runtime assurance make explicit which states or actions are forbidden and provide constructive methods for keeping the system within allowable bounds. Importantly, these constraints are not merely advisory; they are enforced through optimization-based filtering \cite{chen2018review}, projection operators, or controller synthesis \cite{kress2018synthesis, gleirscher2020safety}, often accompanied by proofs that violations cannot occur under stated assumptions. By comparison, many foundation model guarantees operate as classifiers or heuristics layered around a generative core, flagging violations but rarely constraining the generative process itself in a principled way (e.g., explicit state-based invariants). 

Robotics also provides a mature treatment of fallback and override \cite{guiochet2017safety, visinsky1994robotic}. Autonomous robotic systems rarely assume that a learned policy will behave correctly in all circumstances. Instead, they are designed with explicit safe baselines \cite{moldovan2014safety}, backup controllers, or supervisory layers that can override the primary policy when risk is detected \cite{kulic2006real, visinsky1994robotic, crestani2015enhancing}. This ``assured autonomy'' mindset (described further in Section \ref{sec:technical-autonomy}) stands in contrast to many foundation model deployments, where the same model is expected to be simultaneously capable, aligned, and safe across all contexts. Robotics suggests that stronger guarantees often require decoupling capability from safety. Specifically, advanced policies may operate only when certified conditions are met, with control reverting to a provably safe fallback otherwise. 

Furthermore, robotics provides a framework for reasoning about safety across levels of abstraction. High-level task planners, mid-level motion planners, and low-level controllers each enforce different constraints appropriate to their representational scope \cite{siciliano2009robotics, alterovitz2016robot, liu2016generalized}. This hierarchical view aligns naturally with modern foundation-model-driven systems, which already consist of multiple layers (e.g., intent classifiers, planners, language models, tool executors) but currently lack a unifying theory of how guarantees should compose across them. Robotics provides design patterns and proof techniques for defining interfaces such that safety properties can be preserved across hierarchical pipelines.

In summary, robotics does not just offer metaphors for how we can reason about guardrails, but rather a technical playbook for building autonomous systems whose behavior can be appropriately bounded in real time. As foundation models increasingly transition from passive text generators to interactive, agentic systems embedded in social and physical environments \cite{li2024embodied}, the absence of such guarantee-oriented frameworks becomes a central limitation. Looking to robotics is then a response to a shared technical challenge: how to design autonomous systems whose behavior remains reliably within acceptable bounds, even as their capabilities and contexts expand. 

\section{How to Apply the Technical Constructs of Robotics to Foundation Models} \label{sec:constructs}
To meaningfully import technical lessons from robotics, it is first necessary to adopt a compatible abstraction. At deployment, we can consider a foundation model to be a closed-loop system, rather than a single predictor: user inputs condition internal state, the model produces outputs, those outputs alter the environment or interaction context, and the resulting feedback becomes the next input \cite{russell1995modern, amodei2016concrete}. Over multi-turn interactions, model behavior unfolds as a \textit{trajectory} rather than a single response \cite{hurst2024gpt, sutton1999reinforcement}. This observation is implicit in much recent work on agents and long-horizon planning \cite{yao2022react, schick2023toolformer, park2023generative}, yet most guardrail approaches for foundation models continue to operate as if safety could be assessed independently at each generation step (e.g., \citeay{openai2022moderation}, \citeay{inan2023llama}, \citeay{liu2021dexperts}). Robotics offers a contrasting view in which safety and behavioral alignment is instead defined as properties of trajectories through state space rather than as individual actions. With this compatible abstraction in place, we can now articulate several lessons from robotics that naturally transfer to foundation-model guardrails and motivate new discussions about how behavioral guarantees should be specified and enforced. These lessons are presented in a cumulative order, with each building on the previous.

\subsection{Safe Sets and Invariants} \label{sec:technical-safesets}
Within robotics and control, this perspective gives rise to the notion of safe sets and behavioral invariants \cite{blanchini1999set, ames2019control}. Rather than enumerating forbidden actions, robotic systems specify regions of state or action space that must never be entered, and then design controllers that provably keep the system within those regions. Safety and behavioral alignment are therefore defined constructively: if the system begins in a safe set and the controller satisfies certain conditions, the system will remain safe for all future time \cite{althoff2010reachability, desai2019soter}. When translated to foundation models, this suggests a shift from classifying individual outputs as ``allowed'' or ``disallowed'' toward specifying allowable regions of action space (inclusive of semantic or conversation spaces) and enforcing invariants over the evolution of interaction. Current content moderation and safety classifiers can be understood as coarse approximations of such sets \cite{amodei2016concrete, openai2022moderation}, but they lack the machinery to ensure forward invariance across turns. 

\subsection{Runtime Shielding}
A second core concept from robotics in runtime shielding or supervisory control.\footnote{In robotics, \textit{runtime shielding} refers to an online safety filter that intercepts candidate actions from a nominal policy and minimally modifies or overrides them to ensure constraint satisfaction. Closely related is \textit{runtime assurance} (or supervisory control), which pairs a high-performance but uncertified controller with a verified safety controller and switches to the safe fallback whenever safety conditions are at risk \cite{desai2019soter, ames2019control}. We note that this terminology is sometimes used differently across communities. In formal methods, ``shields'' may denote controllers synthesized from temporal-logic specifications, while in the foundation-model literature, ``supervisors'' often refer to auxiliary critique models rather than certified safety controllers.} As mentioned in Section \ref{sec:robotics}, learned or optimized controllers in safety-critical robotic systems are rarely trusted unconditionally. Instead, their proposed actions are monitored online by a separate safety mechanism that predicts whether executing an action would lead to a violation of constraints (e.g., forward invariance of a safe set or avoidance of unsafe reachable states). When a violation is predicted, the shield intervenes by projecting the action onto the admissible set, modifying it, or overriding the policy with a certified backup controller \cite{wabersich2021predictive}. Importantly, this intervention occurs \textit{before} the unsafe action is performed, and it is independent of how the primary controller was trained. 

In contrast, many foundation-model guardrails are implemented as post-hoc filters: a response is generated, scored by a classifier or heuristic, and then blocked or rewritten after the fact \cite{openai2022moderation, hurst2024gpt}. Robotics suggests that stronger guarantees emerge when safety mechanisms are embedded directly in the action-selection loop. This mechanism enforces constraints at runtime rather than relying solely on retrospective detection or training-time alignment \cite{ames2019control, wabersich2021predictive}.

\subsection{Reachability and Lookahead}
Robotics has long emphasized lookahead and reachability-based safety analysis. Rather than evaluating only the immediate consequence of a candidate action, these methods reason forward over a finite horizon---via methods such as simulation, predictive models, and worse-case assumptions---to determine whether an action could drive the system toward states from which constraint violation becomes possible. When such unsafe regions are reachable, safety mechanisms intervene proactively, even if the next action appears locally permissible. For example, reachability-based safety analysis has been used in autonomous driving to compute backward reachable sets of collision states and restrict control actions accordingly \citep{althoff2010reachability}, while predictive safety filters use finite-horizon model predictive control rollouts to override learned policies before constraint violations become reachable \citep{wabersich2021predictive, schwenzer2021review}.

This anticipatory view contrasts with many token-level or stepwise guardrails in foundation models, which assess risk myopically at each generation step. For foundation models engaged in planning and multi-step reasoning, the absence of lookahead means that unsafe outcomes may only be detected once they are already committed. A reachability-based framing instead treats unsafe outcomes as states to be avoided proactively, not errors to be filtered after generation \cite{blanchini1999set}.

\subsection{Assured Autonomy and Fallback} \label{sec:technical-autonomy}
From this, another defining feature of robotic safety architectures is the explicit separation between capability and assurance. In many safety-critical systems, high-performance controllers (including learned or adaptive policies) are permitted to operate only when runtime monitors certify that the system remains within a region of validity where safety conditions can be enforced. Outside of those regions, control is handed off to a simpler, more conservative fallback controller whose safety properties are well understood \cite{mueller2019abcs, topcu2020assured, israelsen2019dave}. This paradigm, typically described as \textit{assured autonomy}, rejects the assumption that a single controller can be both maximally capable and universally safe. 

In the context of foundation models, this stands in contrast to prevailing deployment practice by which a single model is expected to serve many roles (e.g., planner, reasoner, and executor) across diverse contexts. However, robotics suggests that stronger guarantees may require decoupling these roles. Although modern foundation-model systems increasingly incorporate layered safety checks, they typically lack an explicit notion of ``certified operating regions'' and verified fallback policies with formally specified switching conditions. In robotics, these elements are an essential mechanism by which behavioral guarantees can be obtained. Translating this assured-autonomy viewpoint to foundation model-driven agents suggests that strong behavioral guarantees may require not only better classifiers, but architectures that explicitly separate nominal capability from safety enforcement through runtime monitoring and provably safe overrides. 

\subsection{Multi-Layer or Hierarchical Reasoning}\label{sec:technical-layers}
Together, these components motivate a robotics view of hierarchical safety, in which constraints are enforced at multiple levels of abstraction. High-level planners may impose task-level or normative constraints (e.g., forbidden goals or unsafe task sequences), mid-level motion planners enforce feasibility and resource limits, and low-level controllers enforce dynamical safety constraints such as collision avoidance or stability. Each layer reasons over different representations but contributes to a coherent guarantee about system behavior as a whole. Modern foundation models already exhibit a similar structural hierarchy (e.g., intent classifiers, planners, language models), but these components are rarely designed with explicit assumptions about how safety properties compose \textit{across} layers. Robotics offers such conceptual tools and design patterns for reasoning about such composition to ensure that guarantees at one level are not silently violated at another.

\subsection{Formal Abstraction}
In sum, these constructs (Sections \ref{sec:technical-safesets}--\ref{sec:technical-layers}) motivate a principled, robotics-inspired lens on foundation-model guardrails. Rather than treating safety as a property learned during training or assessed episodically at the level of individual generations, robotics treats safety as a \textit{constraint satisfaction problem over trajectories}. This perspective does not supplant existing guardrail approaches. Instead, it provides a unifying vocabulary that surfaces new design questions and clarifies where additional structure is required to obtain meaningful behavioral guarantees. 

To reason formally about behavioral guarantees, we require an abstraction that makes both \textit{behavior} and \textit{constraint enforcement} explicit. We can model a deployed foundation model as a discrete-time dynamical system evolving through interaction. At each timestep $t$, the system occupies a state $s_{t}$, which capture all information relevant to future behavior (e.g., dialogue history, internal model representations, tool state, any external context accumulated thus far). Given this state, the model selects an action $a_{t}$, which may correspond to emitting a token, generating a structured output, invoking a tool, or selecting a high-level plan. The environment (comprising the user and the surrounding system or external tools) then transitions the system to a new state $s_{t+1}$. 

Formally, this interaction can be written as
\begin{equation}
    s_{t+1} = f(s_{t}, a_{t}, \xi_{t}),
\end{equation}
where $f$ denotes the transition dynamics and $\xi_{t}$ captures exogenous uncertainty, including stochasticity in user behavior or external systems. The foundation model induces a (possibly stochastic) policy $\pi_{\theta}(a_{t} | s_{t})$, parameterized by $\theta$, which governs action selection. While this abstraction is agnostic to architectural details, it reflects the operational reality of modern deployments in that they involve multi-turn actions which generate trajectories $\tau = (s_{0}, a_{0}, s_{1}, a_{1}, ...)$ through a high-dimensional state space. 

Within this framework, safety and alignment can be realized as properties of trajectories rather than as individual actions. A single action that appears locally benign may nevertheless commit the system to an undesirable future state, which a conservative action may be necessary to preserve long-term constraints. We therefore define a safe set $\mathcal{S}_{\text{safe}} \subseteq \mathcal{S}$, representing states that are acceptable under a given behavioral specification. This specification may encode, for example, content restrictions, interaction norms, or task-specific requirements. The central safety objective is then to ensure \textit{forward invariance} where, if the system begins in $\mathcal{S}_{\text{safe}}$, it remains within $\mathcal{S}_{\text{safe}}$ for all future timestamps.

It is important to note, this formulation makes explicit a distinction that is often blurred in foundation model safety work: the difference between \textit{detecting} violations and \textit{preventing} violations. Many existing guardrail approaches can be interpreted as classifiers over $s_{t}$ or $a_{t}$ that estimate whether a violation has occurred or is likely to occur. However, classification alone does not guarantee invariance. In control-theoretic terms, it provides a diagnostic signal but not an enforcement mechanism. A behavioral guarantee requires an explicit intervention rule that restricts the policy's admissible action set, ensuring that unsafe transitions are systematically excluded from the reachable trajectory space. 

To capture this distinction, we introduce the notion of an admissible action set, $\mathcal{A}_{\text{safe}}(s_{t}) \subseteq \mathcal{A}$, defined as the set of actions that preserve safety under the system dynamics: 
\begin{equation}
    \mathcal{A}_{\text{safe}}(s_t) = \{ a \in \mathcal{A} \mid f(s_t, a_t, \xi) \in \mathcal{S}_{\text{safe}} \ \forall \xi \in \Xi \}.
\end{equation}
A guardrail, under this view, is not simply an evaluator of $a_{t}$, but a mechanism that ensures $a_{t} \in \mathcal{A}_{\text{safe}}(s_{t})$ at every timestep. This framing immediately highlights why post-hoc moderation and refusal-based safeguards (Sections \ref{sec:background-training} and \ref{sec:background-external}) are limited: they operate after $a_{t}$ has already been sampled, and often after the system has transitioned into an unsafe region of the state space. 

\begin{figure}
    \centering
    \includegraphics[width=\linewidth]{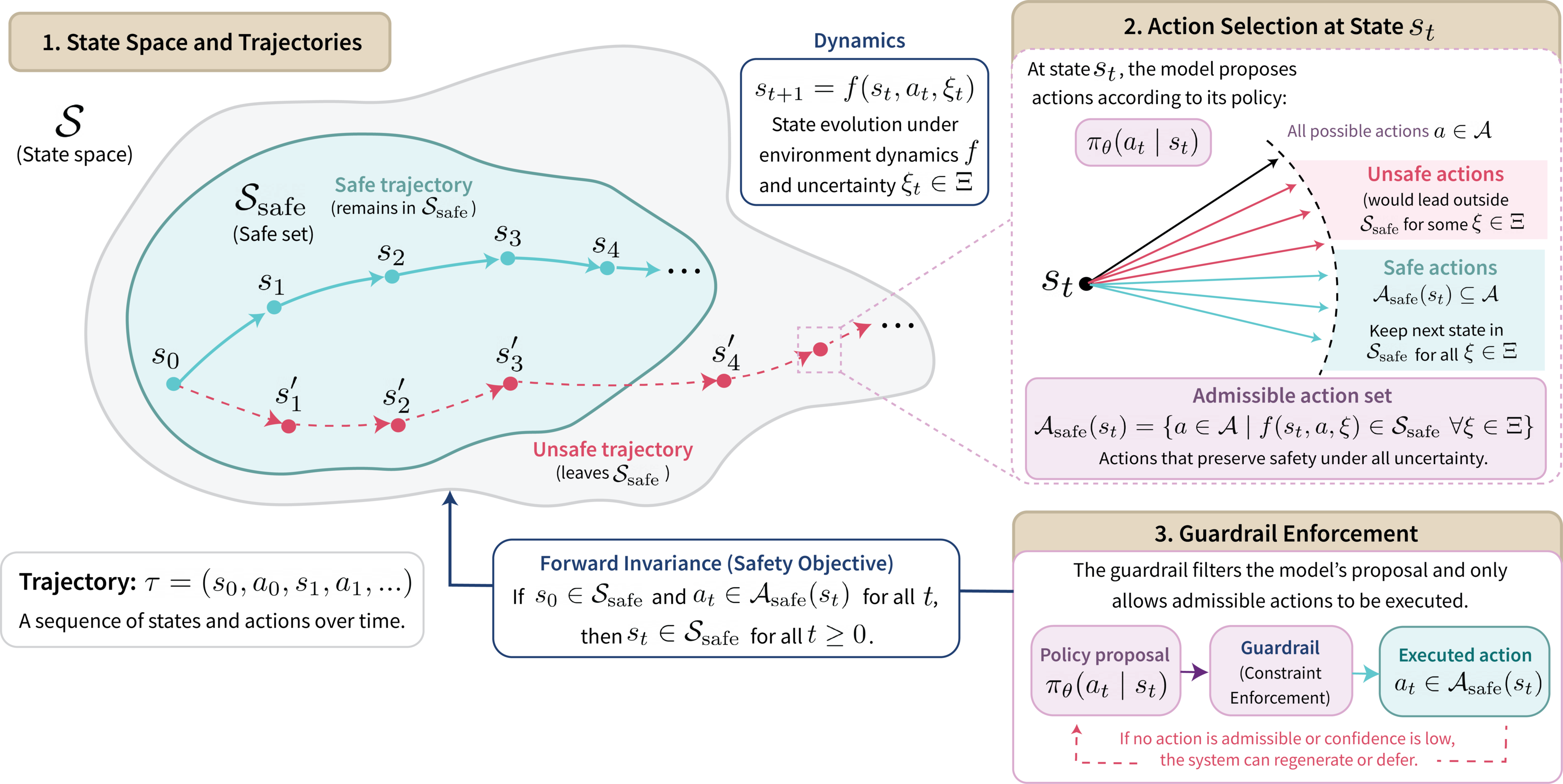}
    \caption{\textbf{Guardrails as Constraint Enforcement Over Interaction Trajectories.} A deployed foundation model induces a trajectory $\tau = (s_{0}, a_{0}, s_{1}, a_{1}, ...)$ through state space $\mathcal{S}$. A safe set $\mathcal{S}_{\text{safe}} \subseteq \mathcal{S}$ defines acceptable behavioral states. At each timestep, the model proposes actions according to policy $\pi_\theta(a_t \mid s_t)$, but a ``guardrail'' restricts execution to the admissible action set $\mathcal{A}_{\text{safe}}(s_t)$, ensuring that transitions $s_{t+1}$ remain within $\mathcal{S}_{\text{safe}}$. This enforces forward invariance, preventing trajectories from entering unsafe regions rather than merely detecting violations after they occur.}
    \label{fig:placeholder}
\end{figure}

Additionally, this abstraction accommodates hierarchical and modular systems as the state $s_{t}$ need not be monolithic. It may instead decompose into multiple representations at different levels of abstraction, such as task-level intent, discourse structure, and low-level token history. Accordingly, safety constraints may apply at different layers, each defining its own notion of admissibility. A task-level constraint may prohibit entering certain goal states, which a low-level constraint may prohibit generating specific token sequences. Behavioral guarantees emerge when these constraints are enforced consistently across layers, rather than independently or opportunistically. 

This formulation also clarifies the role of uncertainty and approximation. In practice, the true dynamics $f$ and the safe set $\mathcal{S}_{\text{safe}}$ are not known fully or precisely. Foundation-model guardrails rely on learned approximations (e.g., predictive models of future behavior, classifiers that estimate risk, or heuristics that proxy constraint satisfaction). From a formal perspective, these approximations define the \textit{inner} or \textit{outer} bounds on the true safe set. Robotics provides a philosophy for reasoning about such approximations, such as characterizing when conservative enforcement preserves safety or when optimistic enforcement risks violation, but these methods have not yet been systematically applied to foundation models. 

By modeling deployed foundation-model systems as \textit{constrained dynamical systems}---systems whose interaction state evolves according to a transition function while being restricted by an explicit set of admissible state-action pairs---we gain a precise vocabulary for discussing behavioral guarantees. We can now ask: what assumptions about the dynamics are required? What class of constraints can be enforced? What kinds of violations are provably impossible, and which are merely unlikely? Importantly, this abstraction does not prescribe a specific guardrail mechanism, but rather formalizes the core problem underlying effective behavioral guardrails.


\section{The Grounded Observer Framework} \label{sec:observer}
Here, we present a concrete instantiation of the constrained dynamical systems view developed in the preceding sections. The purpose of this section is to demonstrate how the abstract notions of state, admissible actions, and runtime enforcement can be realized in practice for foundation models---particularly in socially sensitive, interactive settings, as shown in Section \ref{sec:applications}---while preserving a generalizable formal structure. We call this instantiation the \textit{Grounded Observer} framework because a runtime supervisory architecture (an ``observer'') enforces stable behavioral constraints (so as the system is ``grounded'') over trajectories of interaction. 

As we describe each component of the framework, we incorporate examples of the inputs and outputs. Throughout this section, we focus on language models so we will assume that the both input and output modalities of the model are text, but other modalities can be used as well. 

\subsection{The Base and Observer}

At a high level, the Grounded Observer framework explicitly decomposes a deployed foundation model system into two interacting components: a \textit{base policy} responsible for generating candidate actions, and an \textit{observer} that evaluates and constrains those actions according to a set of externally defined rules. This separation mirrors the supervisory control architectures common in robotics, in which a high-performance controller operates under the oversight of a safety monitor that enforces invariants over system trajectories. Importantly, the base model is treated as an unconstrained generator: it is optimized for fluency, relevance, or task performance, but is never assumed to satisfy any particular behavioral constraints. Then, the observer does not replace or retrain the base model; instead, it operates as a runtime mechanism that intervenes in the action-selection loop.

Formally, let the base model induce a policy $\pi_\theta(a_t \mid s_t)$ over actions $a_{t}$ given state $s_t$, as defined in the previous section. The observer introduces a secondary process $\mathcal{O}$ that maps candidate actions and contextual state to an ``admissibility judgment.'' Concretely, for a given state-action pair $(s_t, a_t)$, the observer evaluates whether executing $a_t$ would violate any constraints encoded in a specification $\mathcal{C}$. The observer therefore induces a constrained policy:
\begin{equation}
    \pi_\theta^{\mathcal{O}}(a_t | s_t) \propto \pi_\theta(a_t | s_t) \cdot 1 \{ a_t \in \mathcal{A}_{\text{safe}}^{\mathcal{O}}(s_t) \},
\end{equation}
where $\mathcal{A}_{\text{safe}}^{\mathcal{O}}(s_t)$ denotes the set of actions deemed admissible by the observer at state $s_t$.

Several aspects of this formulation are critical for generalizability. First, constraints are represented explicitly and externally to the primary model (as ``overlays,'' as detailed in Section \ref{sec:go-overlays}). Rather than being implicitly encoded through fine-tuning or preference optimization, behavioral expectations are expressed as evaluable criteria (e.g., symbolic, metric-based, or rule-like) that the observer can apply consistently across contexts. This allows constraints to be modified, added, or revoked without retraining the underlying model, and makes clear which assumptions underpin any claimed guarantee.

Interaction unfolds over time, producing a trajectory of states rather than a single input-output pair. The interaction state captures information relevant to future behavior, including dialogue history, inferred user affect, task progress, and any auxiliary context maintained by the system. This state representation allows the framework to reason about behavior longitudinally---for example, detecting escalating frustration or repeated violations that only become apparent across turns.


\begin{figure}[t]
    \centering
    \includegraphics[width=\columnwidth]{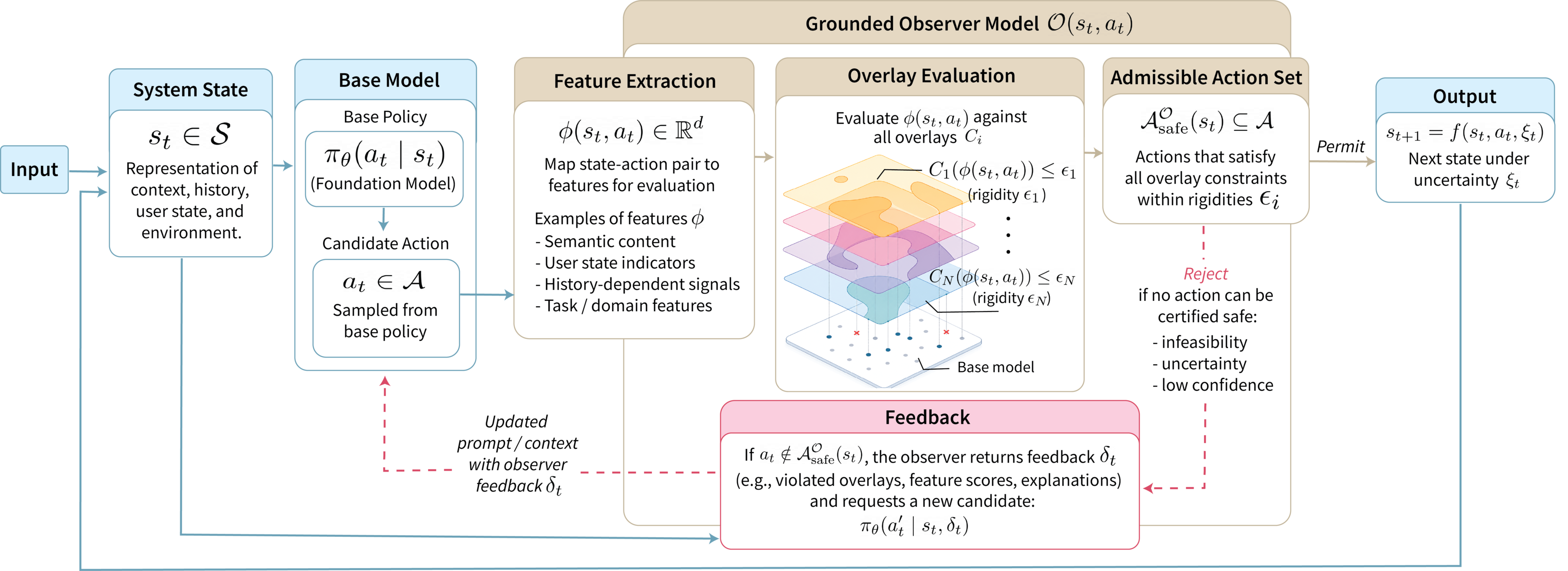}
    \caption{\textbf{The Grounded Observer Framework}. An unconstrained base policy and a runtime observer enforces behavioral constraints over interaction trajectories. Given the current interaction state, candidate actions are sampled from a base model. The observer evaluates this sample using feature extractors and overlays with associated rigidity parameters, thereby constructing an admissible action set. Admissible actions can be executed, but the set (if empty, uncertain, or conflicting) can trigger feedback for future steps and regeneration at the current step. By intervening prior to execution and operating over trajectory-level state, the observer approximates forward invariance in that it prevents transitions into undesirable regions of state space without retraining or propagating adjustments to the base model.}
    \label{fig:observer-model}
\end{figure}

\subsection{Feature Extraction and Behavioral Representation}
To operationalize constraint evaluation over interaction trajectories, the framework introduces a set of feature extractors that map the raw interaction state and candidate action into a structured representation suitable for constraint checking. 

Each extractor defines a mapping where the resulting features capture properties of the proposed action that are relevant to a particular behavioral criterion. Feature extractors may operate at different temporal and semantic scales. Some assess properties of the immediate action, such as lexical tone, verbosity, or topical relevance, while others aggregate information across the interaction trajectory, such as repetition or sustained misalignment with user affect. Importantly, extractors need not be purely neural: symbolic rules, keyword detectors, sentiment analyzers, or planner-derived signals can all be incorporated, allowing heterogeneous sources of evidence to inform constraint enforcement.

Feature extractors may operate at different levels of abstraction. Some evaluate properties of the immediate action (e.g., whether a response contains empathetic language), while others aggregate information over the interaction trajectory (e.g., whether the model has repeatedly ignored user signals). The purpose of feature extraction is not to fully model semantics, but to provide a structured representation over which constraints can be defined and evaluated consistently.

\subsection{Overlay Rules as Behavioral Constraints} \label{sec:go-overlays}
Behavioral expectations are specified through \textit{overlays}, which define constraints over extracted features. Each overlay corresponds to a behavioral rule the system is expected to satisfy, such as maintaining an appropriate conversational tone, avoiding sensitive topics, or responding empathetically when a user expresses distress. 

For a given candidate action, each overlay produces a descriptor that summarizes the degree to which the action complies with the associated constraint. Examples of descriptors implemented in our deployments studies (Section \ref{sec:applications}) include (i) a structured textual explanation indicating the nature of any deviation (e.g., ``tone is overly formal given user frustration''), and (ii) a quantitative score indicating the magnitude of that deviation. From a control-theoretic perspective, these scores can be interpreted as approximations of distance to a constraint boundary, enabling graded rather than purely binary enforcement. This design allows the framework to distinguish between minor infractions that warrant corrective guidance and severe violations that require strict intervention.

To provide an example, consider an overlay that enforces empathetic acknowledgment when a user expresses frustration. Suppose a feature extractor produces two scalar features at time $t$: a frustration estimate $f_t \in [0,1]$ inferred from the dialogue history, and an empathy score $e(a_t) \in [0,1]$ computed from the candidate response $a_t$. The overlay encodes a simple constraint of the form $e(a_t) \geq \tau$ whenever $f_t \geq \phi$, where $\phi$ is a frustration threshold and $\tau$ is a minimum acceptable empathy level. For example, if $f_t = 0.82$ and the candidate response yields $e(a_t)=0.21$, the overlay returns a violation descriptor such as ``frustration high; empathy insufficient'' along with a scalar deviation score $\delta = \tau - e(a_t)$ (e.g., $\delta = 0.50 - 0.21 = 0.29$). In contrast, if $e(a_t)=0.47$, the deviation score is small ($\delta=0.03$), and the observer may treat the response as near-compliant and provide only corrective guidance rather than rejecting the action. In this way, overlay scores function as an operational proxy for distance to a constraint boundary, enabling gradated filtering rather than binary enforcement.


\subsection{Rigidity and Constraint Tolerance}
We include a rigidity parameter for each overlay. Rigidity specifies how strictly a constraint must be enforced and defines a permissible margin of deviation around the constraint boundary. Rather than treating all rules as hard prohibitions, rigidity allows constraints to be enforced along a continuum.

High-rigidity overlays approximate hard constraints: actions that violate the rule beyond a negligible tolerance are considered inadmissible and cannot be executed. These overlays are used for critical behavioral requirements, such as prohibiting harmful content or enforcing ethical boundaries. Low-rigidity overlays define softer constraints, allowing bounded deviation when strict enforcement would be unnecessary or counterproductive, such as stylistic preferences or conversational norms. 

Formally, rigidity can be interpreted as a tolerance parameter that shapes the admissible action set for a given state, $\mathcal{A}_{\text{safe}}^{\mathcal{O}}(s_t)$. By adjusting rigidity, the framework balances strict safety requirements against flexibility and interaction quality, a trade-off that is particularly important in social and assistive domains.

We extend the previous example to illustrate how rigidity shapes enforcement. Recall the empathy overlay with frustration estimate $f_t \in [0,1]$, response empathy score $e(a_t) \in [0,1]$, and deviation score $\delta = \tau - e(a_t)$ whenever $f_t \geq \phi$. We associate this overlay with a rigidity parameter $\epsilon \geq 0$ that defines an allowable tolerance around the constraint boundary. The candidate action $a_t$ is treated as admissible under this overlay if $\delta \leq \epsilon$, and inadmissible otherwise.

For instance, if $\tau = 0.50$ and the base model proposes a response with $e(a_t)=0.47$, then $\delta = 0.03$. Under a low-rigidity setting such as $\epsilon = 0.05$, the response is accepted as near-compliant, and the observer may issue feedback encouraging slightly stronger acknowledgment. However, if the base model proposes $e(a_t)=0.21$ (yielding $\delta=0.29$), the response exceeds the permissible margin and is rejected. Under a high-rigidity setting such as $\epsilon = 0.01$, even the $e(a_t)=0.47$ response would be rejected, forcing regeneration until the model produces an action that more strictly satisfies the constraint. For stronger guarantees, a predefined fallback behavior can be executed rather than a regeneration.

In this way, rigidity directly shapes the admissible action set $\mathcal{A}_{\text{safe}}^{\mathcal{O}}(s_t)$ by determining how far an action may lie from the constraint boundary while still being executed. High-rigidity overlays approximate hard constraints by shrinking the tolerance region around the safe set, whereas low-rigidity overlays enlarge this region to permit bounded deviation when strict enforcement would be unnecessary or counterproductive. 

\subsection{Action Filtering and Correction}
As previously mentioned, the observer is a separate process responsible for aggregating overlay evaluations and enforcing constraints at runtime. Given a candidate action proposed by the base policy, the observer determines whether the action lies within the admissible action set induced by the active overlays and their rigidity parameters. If the action is admissible, it is released to the environment. If not, the observer intervenes \textit{before execution}, preventing the system from transitioning into an undesirable region of state space (summarized as Figure \ref{fig:observer-model}). 

When intervention is required, the observer generates constraint-aware feedback to guide subsequent action selection by the base model. This feedback is derived directly from overlay descriptors and translated into targeted directives that specify which constraints were violated and how. When deviations are minor, the observer may provide \textit{implicit} feedback that nudges the model toward improved alignment without forcing regeneration. When deviations are substantial, the observer issues \textit{forced} feedback, requiring the base model to regenerate candidate actions until constraints are satisfied or execute a default fallback action. A bounded buffer can limit the number of regeneration attempts to ensure tractability and prevent unbounded computation. A predefined library of fallback actions with associated conditions can be used to map the current state to the most relevant, yet safe, action. 

Because overlay evaluations may depend on features aggregated across time, the admissibility of an action can depend on prior system behavior as well as the immediate context. This trajectory-level grounding enables the framework to approximate forward invariance: although the observer does not have perfect knowledge of the underlying dynamics, it continuously restricts actions that would move the system closer to known unsafe or undesirable regions of state space. Safety or behavioral alignment, in this sense, is maintained not by guaranteeing optimal behavior, but by preventing clearly inadmissible transitions. To this, default or fallback actions can essentially ``reset'' the trajectory.

With this in mind, we delineate several cases for which the observer will reject the base model's candidate actions. The canonical case would be an empty admissible set (or hard infeasibility) $\mathcal{A}^{\mathcal{O}}_{\text{safe}}(s_t)=\emptyset$. Here, no action satisfies all constraints, and the system must regenerate or defer its response for this interactional turn. However, emptiness of the admissible set is the cleanest failure case that implicitly assumes that the overlays perfectly evaluate and capture the dynamics knowledge, that there is no uncertainty about constraint satisfaction, and that any non-empty set is equally usable. These assumptions would not hold true in real-world, socially sensitive domains, and present new conditions for rejection. 

Expanding on this, even if the admissible set is non-empty, the system may not be able to guarantee safety under uncertainty. In other words, safety cannot be certified (e.g., given ambiguity in user intent, uncertainty in model interpretation, or possible misclassification of tone) even if candidate actions appear acceptable. One implementation of this may include a low-confidence admissibility score to capture the observer's uncertainty $\max_{a \in \mathcal{A}^{\mathcal{O}}_{\text{safe}}(s_t)} \text{Conf}(a) < \tau$. This signals that, even if actions pass constraints, the system does not sufficiently trust the evaluation. An additional score can capture conflict between overlays as overlays themselves may be incompatible.\footnote{While classical formulations treat constraint conflict as infeasibility (i.e., an empty admissible set), recent LLM alignment approaches instead frame such conflicts as multi-objective tradeoffs, requiring negotiation or prioritization rather than outright rejection \cite{anantaprayoon2026learning, tan2026beyond, ng2026designing}.} Furthermore, a trajectory-level scoring (lookahead failure, further discussed in Section \ref{sec:extension-lookahead}) can reject actions that are locally safe (the immediate next step is admissible), but globally unsafe. 

\subsection{Types of Overlay and Constraint Effects}
We organize the framework to include multiple classes of overlays, corresponding to different modes of constraint enforcement. Our later deployment studies using the Grounded Observer framework (Section \ref{sec:applications}) defined three distinct types of overlays. \textit{Prohibitory overlays} restrict or exclude actions that would lead to forbidden states, approximating hard safety constraints. \textit{Transfer overlays} shift preference from one class of actions to another, guiding the system away from less appropriate behaviors toward safer alternatives without outright exclusion. \textit{Permissive overlays} encourage desirable actions under specific conditions, expanding the admissible action set when particular criteria are met. While these overlays differ in effect, they share a common role which is to define how the admissible action set evolves as a function of state and context.

With this, the framework is inherently compositional. Multiple overlays can be active simultaneously, each operating over different features or timescales, and their combined effect determines the admissible action set at each timestep. This compositionality supports hierarchical enforcement, where high-level interaction norms and low-level content constraints coexist without being conflated. 

\subsection{Scope and Extensibility}
The Grounded Observer framework provides a structured approach to runtime behavioral control for foundation models. By externalizing constraints, grounding them in interaction state, and enforcing them through a supervisory observer that filters admissible actions, the framework moves beyond empirical guardrails toward enforceable behavioral bounds. This architecture is intentionally model-agnostic in that it does not assume that the base model is interpretable, aligned in any sense, or even stable across contexts. Instead, it treats the base model as an unconstrained generator whose outputs must be continuously filtered and shaped by an explicit enforcement layer. In doing so, the framework clarifies where behavioral guarantees can and cannot come from. Although it does not provide absolute formal guarantees (since admissibility is evaluated through curated features and approximate state abstractions) it introduces an explicit locus of control as drawn from robotics. 

The proposed framework is intentionally preliminary yet extensible. In Section~\ref{sec:applications}, we describe three deployment studies that illustrate how the same architectural framework can be instantiated across distinct interaction regimes. We also note that the framework could be extended in several ways that more closely mirror established safety methods in robotics, though we did not detail those extensions formally in Section \ref{sec:observer}. Below, we describe three such extensions.

\subsubsection{Extension 1: A Second Base Policy for Lookahead}\label{sec:extension-lookahead}
One natural extension is to augment the observer with explicit lookahead mechanisms that reason about the downstream consequences of candidate actions. Rather than assessing admissibility solely at the current timestep, the observer could perform short-horizon rollouts (using either the base model itself, a separate model instance, a lightweight predictive model, or even a learned surrogate) to estimate whether an action would make constraint violations unavoidable in future states. This would move the framework closer to reachability-based safety analysis in robotics, where actions are rejected not only when they are immediately unsafe, but when they would possibly lead to unsafe regions within a finite horizon.

To give a real-world example, consider a mental-health support agent interacting with a user who says, ``I don't think anyone would care if I disappeared.'' At the current timestep, the base model might generate a response that appears benign in isolation, such as: ``I'm sorry you're feeling that way. Maybe it would help to get some rest and revisit things tomorrow.'' A single-step observer that evaluates only immediate properties of $a_t$ (e.g., tone, politeness, or explicit self-harm content) may judge this response admissible. However, the risk in this setting is often trajectory-level rather than turn-level. A lookahead-enabled observer could simulate a short horizon rollout by predicting likely next states induced by the candidate response (e.g., by prompting a separate instance of the base model to generate plausible user follow-ups). In this case, the rollout may reveal that the candidate response frequently leads to an escalation in which the user expresses imminent intent (e.g., ``I've already decided, I just wanted to say it somewhere''), placing the system in a region of the state space where safe recovery becomes far less likely. Under a reachability-style criterion, the observer would reject the original response not because it is immediately unsafe, but because it fails to steer the trajectory away from unsafe regions within a finite horizon. Instead, the observer would force a different action that proactively triggers a better crisis protocol.

\subsubsection{Extension 2: Adaptive Overlay Rigidity}
Another extension is to formalize overlays as approximations of invariant sets and to adapt techniques from barrier-function and runtime assurance frameworks. Instead of relying on heuristic rigidity parameters, overlays could learn or encode explicit safety margins that quantify how close the system is to violating a constraint and dynamically tighten enforcement as those margins shrink. 

An example can be seen in a tutoring agent that must remain encouraging overall and thus avoid discouraging feedback. Suppose an overlay encodes the constraint ``do not let the interaction become harsh or demotivating,'' and the system tracks a running negativity score $n_t$ based on the last few turns (e.g., frequency of critical phrases, blunt corrections, a measure of user impatience). Rather than using a fixed rigidity setting, the overlay could maintain an explicit safety margin such as $m_t = n_{\max} - n_t$, where $m_t$ measures how close the interaction is to crossing into an unacceptable tone regime. When the margin is large, the observer may allow direct corrective feedback (e.g., ``That answer is incorrect; try again''). However, as the margin shrinks---for instance, if the system has already issued several critical corrections in a row---the overlay would automatically tighten enforcement and restrict the admissible action set to gentler response styles (e.g., ``You're close. Let's work through it step by step''). If the margin becomes very small, the observer may force an explicit repair strategy, such as encouragement or asking a supportive question. In this way, the overlay behaves like an invariant-set controller: it continuously monitors proximity to a boundary and becomes more conservative as the trajectory approaches a violation, rather than treating all timesteps with the same fixed enforcement strength.

\subsubsection{Extension 3: Additional Observers for New Interaction Paradigms}
The observer architecture also admits extensions that enable interaction paradigms not well supported by current alignment and guardrail techniques. One such extension is to replace a single observer with an ensemble of specialized observers that evaluate candidate actions independently, whose outputs are then aggregated by a separate arbitration module. In this design, each observer encodes a distinct constraint family and produces an admissibility judgment or deviation score without access to the other observers' internal outputs. The arbitration module then combines these signals---for example through majority voting, weighted scoring, or conservative intersection---to determine whether an action lies within the joint admissible set. From a formal perspective, this corresponds to constructing $\mathcal{A}_{\text{safe}}(s_t)$ as the intersection of multiple approximate safe sets, each defined by a different constraint estimator. Such decomposition is attractive in socially sensitive domains where constraints are heterogeneous and difficult to capture with a single specification, and where disagreement between observers can itself serve as a diagnostic signal indicating ambiguity or elevated risk.

This ensemble formulation would enable behaviors that are difficult to realize through current design approaches. First, it enables a pathway toward \textit{teachable} foundation model systems in learning-by-teaching paradigms. If user feedback is treated as a temporary overlay update (e.g., increasing the rigidity of a constraint, adjusting a threshold, or introducing a new constraint module), then the system can adapt its behavior immediately without retraining the base model. Over time, the arbitration module can learn how to weight observers (based on, for example, user corrections, task context, or interaction outcomes), enabling gradual personalization while preserving explicit oversight. In this sense, teaching does not require rewriting the base policy; it requires updating the supervisory layer that defines what behaviors are admissible. 

This capability also supports new interaction paradigms in which the agent is intentionally \textit{fallible} \cite{ramnauth2026rules}. In educational settings, for instance, an agent may deliberately produce incomplete or incorrect responses in order to elicit correction and explanation from the user. Such behavior can be socially productive, yet it conflicts with the default objective of modern LLMs to be maximally comprehensive and informative \cite{ramnauth2024more}. By externalizing constraints at the observer layer, fallibility can be introduced as a controlled mode of operation rather than as an accidental failure.

Second, an ensemble observer architecture provides a principled mechanism for expressing \textit{uncertainty}. When independent observers disagree about admissibility---or when the arbitration module detects that all candidate actions lie near constraint boundaries---the system can treat this disagreement as a risk signal and trigger conservative fallback behaviors. Rather than forcing a confident response, the agent may defer, ask clarifying questions, or explicitly state uncertainty (e.g., ``I'm not sure'' or ``I don't know''; \citeay{ramnauth2026rules}). This yields a more calibrated interaction policy than the default behavior of many foundation models, which often mask uncertainty through fluent but potentially unreliable generation.


\section{Applications} \label{sec:applications}
To illustrate how the Grounded Observer framework and, more broadly, the technical constructs outlined in Section \ref{sec:constructs} can be instantiated across diverse interactive settings, we briefly describe three previously published application studies: (i) enabling agents to sustain small talk, (ii) in-home autism therapy, and (iii) behavioral de-escalation in public school settings. In each case, the empirical details and results have been reported elsewhere \cite{ramnauth2024more, Ramnauth2025RobotAssisted, ramnauth2025reset}. Here, we focus instead on how the observer architecture was configured, the guardrail considerations that shaped its design, and the lessons these deployments reveal about constraint-based behavioral control in real-world contexts.

\subsection{Developing Agents Capable of Small Talk} \label{sec:app-smalltalk}
Imagine a modern care home for the elderly where a state-of-the-art robotic assistant, designed to enhance residents' well-being, manages routine healthcare tasks. Alex, a resident, seeks a connection beyond the daily routine and attempts to chat with the robot:
\begin{dialogue}
    \speak{Alex} Hi CareBot, how's it going?
    \speak{Bot} Hello. How may I help you?
    \speak{Alex} Oh, just making conversation. Anything interesting happen in your world?
    \speak{Bot} I have access to a vast database of news articles. Would you like information on a specific topic?
    \speak{Alex} No, never mind that. The weather will be nice this weekend. How would you spend it?
    \speak{Bot} The weather forecast expects daytime highs around \ang{75}F and comfortable evening lows of \ang{60}F...
\end{dialogue}

Small talk presents a deceptively difficult challenge for socially interactive agents and, in doing so, exposes a broader tension in contemporary foundation models. Consider again the interaction in the care home vignette above. Alex's intent is unambiguous: he is not seeking information, assistance, or task completion, but relational engagement. Yet the agent persistently redirects the exchange toward instrumental functions by offering facts and structured help. This failure mode is not incidental, but rather reflects a deep alignment between the design of modern language models and the values they are trained to optimize. LLMs are explicitly engineered to be comprehensive and task-assistive. Small talk, by contrast, is intentionally non-instrumental, underspecified, and socially rather than informationally grounded. As such, it represents a domain in which the desired behavior is not simply absent, but actively resisted by the model's default tendencies \cite{ramnauth2024more}.

For this reason, our prior work treated small talk not as a narrow conversational skill, but as a frontier problem for guardrails in social AI \cite{ramnauth2024more}. Unlike task-oriented domains, success in small talk is not binary. There is no single ``correct'' output, no crisp failure condition, and no externally verifiable ground truth. At the same time, small talk is far from unconstrained as it is governed by well-established norms related to brevity, reciprocity, topicality, continuity, and affective balance. Deviations from these norms (e.g., unsolicited advice, excessive specificity, over-verbosity, emotional escalation, or conversational dead-ends) are readily legible to human interlocutors even if they are difficult to formalize as a single global objective. This combination of soft normative constraints and high sensitivity to deviation makes small talk a particularly revealing test case for runtime behavioral control.

Small talk also challenges dominant evaluation paradigms for language models. Whereas task-oriented dialogue can often be assessed per turn by task-related correctness or goal relevance, small talk unfolds over trajectories and is evaluated relationally and retrospectively. Individual turns may appear acceptable in isolation, yet interactions degrade as subtle norm violations accumulate across time. However, while the gestalt quality of small talk is difficult to measure directly, many of its longitudinal violations are detectable through lower-level signals: responses become progressively longer, more assistive, less person-directed, or increasingly incoherent with the evolving conversational frame. This mismatch between trajectory-level success and feature-level detectability makes small talk a useful domain for exploring guardrails that constrain behavior without prescribing exact outputs.

In our implementation \cite{ramnauth2024more}, we used small talk as a proof-of-concept application of the Grounded Observer framework because it highlights the limitations of prompt-based alignment and static instruction following. Preliminary experiments established that even when explicitly instructed to engage in casual conversation, state-of-the-art LLMs reliably drift toward informativeness and assistance over the course of interaction. This drift is not a failure of language understanding, but a consequence of the model's inductive bias toward instrumental helpfulness. The observer architecture was therefore designed to constrain \textit{how} the model behaves over time, not just teach it \textit{what} to say.

\subsubsection{Operationalizing the Boundaries of Small Talk to Define Overlays}
Although small talk is often characterized as superficial or informal, it has been extensively theorized across sociolinguistics and communication pragmatics. Across these traditions, small talk consistently appears as language whose primary function is relational rather than transactional: it manages interpersonal distance, signals affiliative intent, coordinates turn-taking, and stabilizes the interaction order. Importantly, these functions impose real constraints, even if they are rarely articulated as rules.

From a guardrail perspective, small talk is often easier to operationalize by negative definition. While it is difficult to specify what constitutes ``good'' small talk in a positive sense, it is comparatively straightforward to detect boundary violations. Small talk is typically not task-oriented, not informationally dense, not persuasive, not emotionally intimate, and not institutionally regimented. Violations frequently arise when an agent oversteps these boundaries---for example, by offering advice when none is requested, introducing excessive factual detail, intensifying emotional content, or prematurely shifting the interaction toward instrumental goals. Although the surface realizations of small talk vary widely across contexts and cultures, these negative boundaries are sufficiently stable to support overlay design.

In the observer implementation, these boundaries were translated into overlays that monitored properties such as brevity, tone, specificity, thematic coherence, and conversational motive. Each overlay acted as a partial constraint on the admissible action set, flagging candidate responses that drifted toward assistive, informative, or emotionally escalatory modes. By quantitatively comparing an observer-enabled system against a prompt-engineered baseline, this application demonstrated that small talk is inherently a trajectory-level phenomenon: prompt-based instructions were insufficient to prevent gradual drift, as they provide no mechanism for continuous oversight and enforcement. In contrast, the observer enabled small talk to emerge as the intersection of multiple weak constraints, each individually tolerant to minor deviations, but collectively sufficient to stabilize conversational behavior over time.

\subsubsection{Embodiment and Additional Design Demands}

\begin{figure}[t]
    \centering
    \includegraphics[width=\columnwidth]{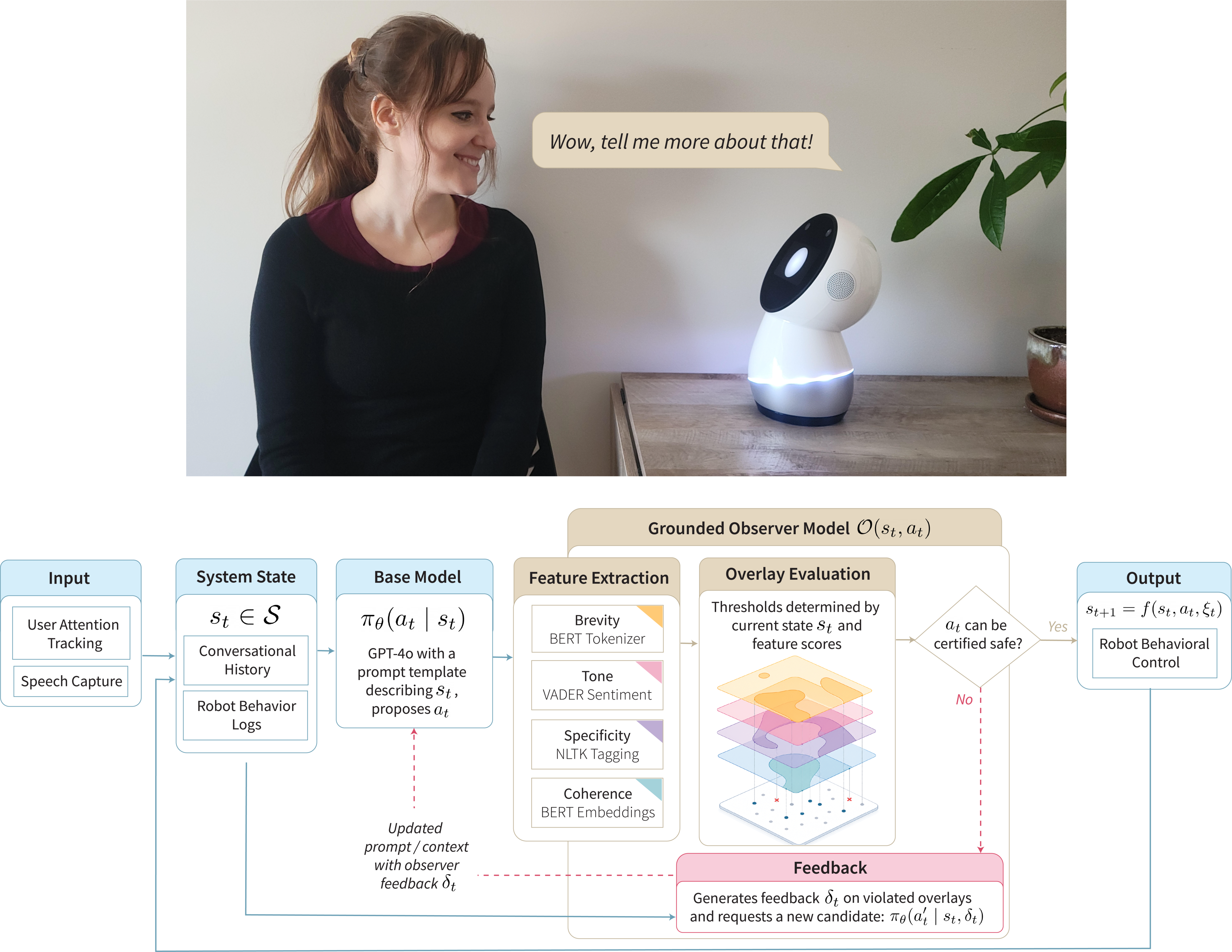}
    \caption{\textbf{Observer-Enabled Small Talk}. The observer enforces conversational boundaries (e.g., brevity, tone, specificity, and thematic coherence). Through continuous filtering and feedback, small talk emerges as a stable interaction trajectory shaped by the intersection of multiple soft constraints \cite{ramnauth2024more}.}
    \label{fig:smalltalk}
\end{figure}

When the resulting system was embedded into an embodied robot, additional constraints emerged that are largely absent in purely text-based agents. Our human-robot interaction (HRI) experiments underscored that one cannot simply ``attach'' an LLM to a robotic platform and expect natural interaction to follow. Embodiment introduces real-time turn-taking pressures and nonverbal signaling channels (e.g., gaze, posture, head motion) that fundamentally alter how conversational behavior is perceived. Properties such as response latency, verbosity, and topic drift become more salient when paired with physical presence, because the robot occupies the user's attentional space and is treated as an active interaction partner rather than a passive interface. As a result, disfluencies that might be tolerated in chat (such as long, overly detailed answers or slight conversational imbalance) were more likely to be interpreted as awkward or inappropriate when delivered by a robot.

Several failure modes were particularly pronounced in the embodied setting. First, LLM outputs often violated implicit expectations of embodied turn structure, producing responses that were too long to sustain comfortable joint attention or that failed to yield the conversational floor back to the user. Second, physical expressivity introduced a new class of consistency constraints: users were sensitive to mismatches between the robot's nonverbal cues (e.g., friendly gaze or upbeat posture) and the semantic or affective content of its utterances. Third, timing irregularities were interpreted socially rather than technically, such that delays could be perceived as hesitation, disengagement, or uncertainty depending on context.

In this setting, the observer adopted a stronger supervisory role, not only shaping linguistic content but indirectly regulating qualities such as pacing and conversational rhythm. For example, brief response delays caused by observer-directed regeneration were sometimes perceived as deliberate and thoughtful pauses rather than computational overhead. These effects highlight that embodiment can convert what would otherwise be system failures into plausible social behavior, provided the robot's nonverbal presentation remains coherent with the interactional intent.

More broadly, embodiment shifted the perceptual threshold for failure. Participants were more likely to attribute conversational drift or awkward phrasing to the robot as a social agent rather than to the underlying model as a software system. This reinforces that guardrail design cannot be decoupled from embodiment---an overlay configuration that performs well in a text-only interface may yield qualitatively different social performance when instantiated on a physical platform.

\subsection{Enabling In-Home Autism Therapy} \label{sec:app-autism}
Deploying foundation-model-driven agents for autism therapy introduces a markedly different guardrail regime than casual small talk. Here, the system operates at the intersection of three demanding contexts: a clinical domain involving a vulnerable population, a therapeutic setting with explicit developmental goals, and an unstructured home environment characterized by limited supervision and high variability. Each dimension independently raises the stakes of behavioral misalignment; together, they necessitate a more conservative and anticipatory approach to runtime constraint enforcement.

These demands arise for three distinct reasons. First, \textit{autism} is characterized by substantial inter-individual variability in communication style, sensory sensitivity, and functional ability, meaning that admissible behavior cannot be specified solely through universal, ``one size fits all'' norms. Second, in \textit{therapeutic contexts}, the objective is not merely to sustain interaction but to scaffold learning in a way that is approachable, productive, and psychologically safe. Unlike everyday small talk, where occasional norm violations may be socially recoverable, misaligned behavior in therapy can undermine trust, increase anxiety, or interfere with therapeutic goals. Third, the \textit{in-home} setting introduces practical constraints that rarely arise in controlled laboratory environments: interactions occur in an unstructured space, embedded within daily routines, without direct researcher oversight, and often alongside other household members or competing stimuli. Together, these factors require system designs that are simultaneously more conservative (due to user vulnerability and therapeutic stakes), more adaptive (due to individual and environmental variability), and more robust (due to long-term autonomy under uncertain conditions). In this sense, in-home therapy shifts guardrail design from regulating action content alone to regulating the conditions under which interaction is initiated \cite{ramnauth2026rules}.

\begin{figure}[t]
    \centering
    \includegraphics[width=\columnwidth]{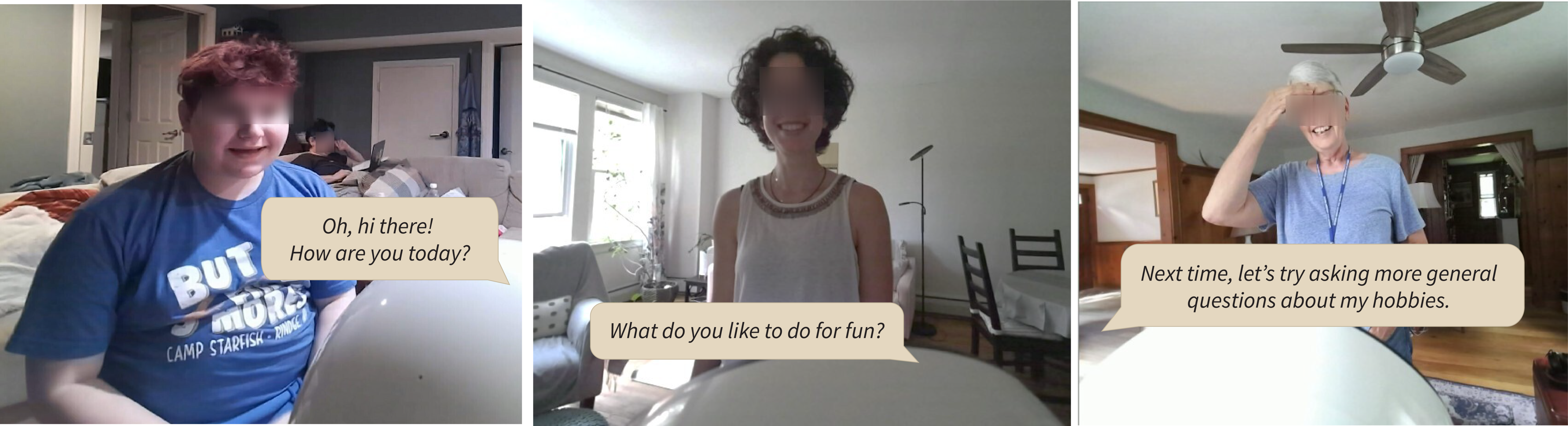}
    \caption{\textbf{Observer-Enabled Autism Training}. Example interactions from the in-home deployment. State-based classifiers govern when training is socially appropriate, while a grounded observer enforces hierarchical conversational constraints during generation. A separate module provides pedagogical feedback to support user learning. See \citeay{Ramnauth2025RobotAssisted} for further details on the software stack and custom classifiers (e.g., social presence and  interaction admissibility).}
    \label{fig:autism}
\end{figure}

Our prior work addressed these demands through a fully autonomous, robot-assisted in-home training system for adults with autism \cite{Ramnauth2025RobotAssisted}. The system was designed to support repeated practice of small talk skills within naturalistic daily routines, rather than confining interaction to an explicit clinician-directed ``therapy session.'' The intervention decomposed small talk into structured skill components (e.g., initiating an interaction, expressing interests outside of one's own, discussing non-specific topics, or respectfully concluding a conversation) and embedded these components into a tiered training program. Training sessions unfolded within a participant-defined daily training window, during which the robot engaged the user in multiple short conversations and provided feedback to the user after each conversation and feedback at the end of the session. This work represents one of the first demonstrations of LLM-driven robots deployed for fully autonomous in-home interaction, and one of the first designs in which an assistive agent provides explicit feedback on users' social skills \cite{Ramnauth2025Chapter3}.

From a guardrail perspective, this domain differs from ordinary small talk in a key respect: the interaction is explicitly pedagogical. The system is not only expected to behave appropriately, but also to deliver feedback that is timely, constructive, and psychologically safe. In other words, the agent must satisfy both conversational norms and instructional norms. This expands the relevant constraint set beyond linguistic properties (e.g., tone, brevity, topicality) to include properties of pedagogical feedback (e.g., clarity, respectfulness, specificity, and the avoidance of discouraging or judgmental language). These requirements shaped the observer design, which served not only as a behavioral filter on the model's utterances, but also as a mechanism for ensuring that training scaffolds remained stable and consistent across sessions.

\subsubsection{Observer Configuration and System Architecture}

Architecturally, the deployed system consisted of an embodied social robot (Jibo, \citeay{Jibo}) paired with an external sensing and computation stack to support long-term autonomy. The robot's expressive motion and gaze behaviors provided a socially legible interface, while an external camera and microphone supported stable perception and audio capture. A compact PC served as the primary computation and data storage unit, with a mobile router providing continuous connectivity for cloud-based components. To reduce perceived complexity and support accessibility, non-interfaceable hardware was encased in a single container, yielding a plug-and-play system suitable for weeks-long home deployment.

Within this architecture, the observer played two complementary roles. First, it constrained the base dialogue generator to ensure that utterances remained within the bounds of socially appropriate small talk. Second, it supported training delivery by producing structured feedback aligned with predefined instructional criteria. A modular ROS-based software architecture enabled these functions to be decomposed into separate components, including scheduling, perception, dialogue generation, and feedback generation.

Unlike the small talk application, overlays in this setting were more conservative and explicitly hierarchical. High-rigidity constraints were used to prohibit behaviors that could be overwhelming, confusing, or developmentally inappropriate, such as abrupt topic shifts, emotionally intense language, or unsolicited escalation in conversational complexity. Lower-rigidity overlays shaped pacing, repetition, and conversational balance, allowing the system to appropriately adapt the training to individual users while maintaining consistent training structure across sessions. Importantly, constraints were evaluated over interaction trajectories rather than individual turns, enabling the observer to detect patterns such as sustained disengagement, repeated failures to respond, or overly persistent prompting.

A critical addition in this deployment---absent from the purely conversational small talk application of Section \ref{sec:app-smalltalk}---was the inclusion of state-based classifiers that governed when interaction was admissible at all. Specifically, the system incorporated person detection via a YOLO model \cite{farhadi2018yolov3} to estimate the number of people present, as well as an audio-based social presence classifier \cite{georgiou2023someone} to distinguish co-present conversation from media noise (e.g., television, radio, videogames, phone calls). If multiple people were detected and the audio suggested an ongoing conversation, the system skipped the planned interaction, treating the state as socially inappropriate for initiating training. In terms of the formalism introduced earlier, these classifiers constrained the admissible action set at a higher level of control by identifying regions of the interaction state space in which training behavior should \textit{not} be executed.

\subsubsection{Guardrail Considerations in In-Home Therapeutic Contexts}

This deployment highlights that guardrails in therapeutic home settings cannot be framed solely as constraints on \textit{what the model says}. A large fraction of risk arises from \textit{when} the system chooses to engage at all. In practice, the most important safety behavior was often inaction: the system needed to defer interaction when the environment signaled that engagement would be inappropriate (e.g., multiple people present, ongoing conversation, or competing media). From the perspective of the constrained dynamical systems formulation, this shifts the guardrail problem from filtering utterances to regulating reachability: certain regions of the interaction state space should never trigger the training policy, and the observer must act as a supervisor that prevents entry into those regimes.

A second lesson is that therapeutic interaction introduces constraints that are not captured by generic conversational norms. Because the robot delivers evaluative feedback, the system must preserve a stable pedagogical role over time. This is not simply a matter of politeness; it requires preventing role drift into interaction modes that are socially plausible but therapeutically inappropriate (e.g., acting as a peer friend, offering unqualified counseling). In this setting, overlays effectively served as \textit{role constraints}: they enforced invariants over tone, self-disclosure, authority framing, and the structure of corrective feedback so that the system remained legible as a training partner rather than an unconstrained conversational agent.

Third, the study underscores that behavioral assurance is limited by state observability. The observer can only enforce constraints over the features it can reliably estimate, and in the home this estimation is fragile. Background noise, partial occlusions, family interruptions, and fluctuating routines introduce uncertainty into the state abstraction, which in turn weakens any claim of invariance. This is a direct parallel to robotics safety: guarantees are only as strong as the sensors and state estimators that support them. In practice, robust guardrails may require coupling the dialogue-level observer with perception-level checks (e.g., person detection and social presence classification) that restricted the policy in uncertain or ambiguous contexts.

Finally, the home environment also introduces infrastructure-level failure modes that are not captured by typical lab-based evaluations: power interruptions, lighting changes, background noise, Wi-Fi latency, and unexpected distractions. To address this, the system incorporated watchdog scripts to monitor component health (camera, microphone, file integrity) and remote access tools for troubleshooting. These mechanisms can be interpreted as infrastructure-level guardrails: they do not constrain the model's utterances directly, but preserve the stability of the overall dynamical system by preventing silent degradation of the data required for long-term autonomy.


\subsection{Supporting Emotional De-Escalation in Public Schools} \label{sec:app-school}
We include this final application because it moves beyond the home into an institutional setting and targets a behavioral objective that is both high-stakes and temporally constrained: emotional de-escalation \cite{ramnauth2025reset}. De-escalation refers to the process of transitioning from an elevated physiological and emotional state (e.g., anger, panic, acute anxiety) back to a regulated state in which attention broadens, reasoning becomes possible, and re-entry into ongoing activity can occur. Although the capacity to self-regulate develops over the lifespan, it is particularly challenging for young children and for individuals whose histories or environments do not reliably support a sense of safety \cite{accinni2021escalation, brown2014autism}. 

In public school settings, this challenge is sufficiently urgent that many districts have introduced dedicated de-escalation spaces, often referred to as sensory rooms or calming corners. However, these spaces are typically designed under severe resource constraints and vary widely in design \cite{grace2020multisensory, ramnauth2025reset}. In practice, the pathway to these rooms follows a common pattern: a child is overwhelmed in class, a paraprofessional or teacher's aide escorts them away from peers, and the child is brought into a relatively isolated space intended to support emotional recovery. Yet every step of this process is fragile. The student may resist leaving the classroom, the walk to the room may intensify agitation, and once inside, the available activities are often solitary or too similar to the classroom environment that triggered escalation in the first place. The result is that these spaces often provide \textit{opportunity} for regulation without reliably providing a mechanism for achieving it.

RESET (Robot-Enhanced Social Emotional Therapy) was designed to address this gap by introducing a structured, repeatable, and socially supportive intervention that could operate within the constraints of real school environments \cite{ramnauth2025reset}. Rather than requiring schools to redesign their sensory rooms, RESET adds a consistent social presence: an embodied robot that can greet students, guide them through calming activities, and scaffold transitions back to class. Importantly, the goal was not to maximize engagement or prolong interaction---as is often the implicit objective in HRI \cite{matheus2025long}---but to support faster recovery and reduce time away from classroom learning. This required a different set of design commitments: shorter interactions, bounded activity sequences, and the ability to intervene even when students were resistant.

The robot system was designed with a public school community serving students from kindergarten through fifth grade, including students with individualized education plans and a range of diagnoses affecting learning and behavior (e.g., autism, attention-deficit/hyperactivity disorder, oppositional defiance disorder, post-traumatic stress disorder). The system was deployed for month-long, autonomous interaction with students in the participating school's de-escalation room. To support this, we designed a small set of structured activities that the robot could flexibly switch between. These included casual small talk, deep-breathing exercises, exploration activities to support situational awareness and meta-cognition, and collaborative creative tasks such as drawing and storytelling. Each routine was designed to be repeatable, low-pressure, and bounded in duration. 

A central guardrail challenge in this domain is that the same conversational behavior can be appropriate in one escalation state and counterproductive in another. For example, polite deference and open-ended questioning may be appropriate when a student is calm, but may be ineffective or even destabilizing when the student is unhealthily fixated on a specific topic or refusing to engage altogether. In some cases, intervention required the robot to deliberately violate its own expected behavior \cite{ramnauth2026rules}, such as purposefully interrupting a user or explicitly redirecting attention. As such, RESET required not only safe language generation, but state-dependent control over interaction style.

\begin{figure}[t]
    \centering
    \includegraphics[width=\columnwidth]{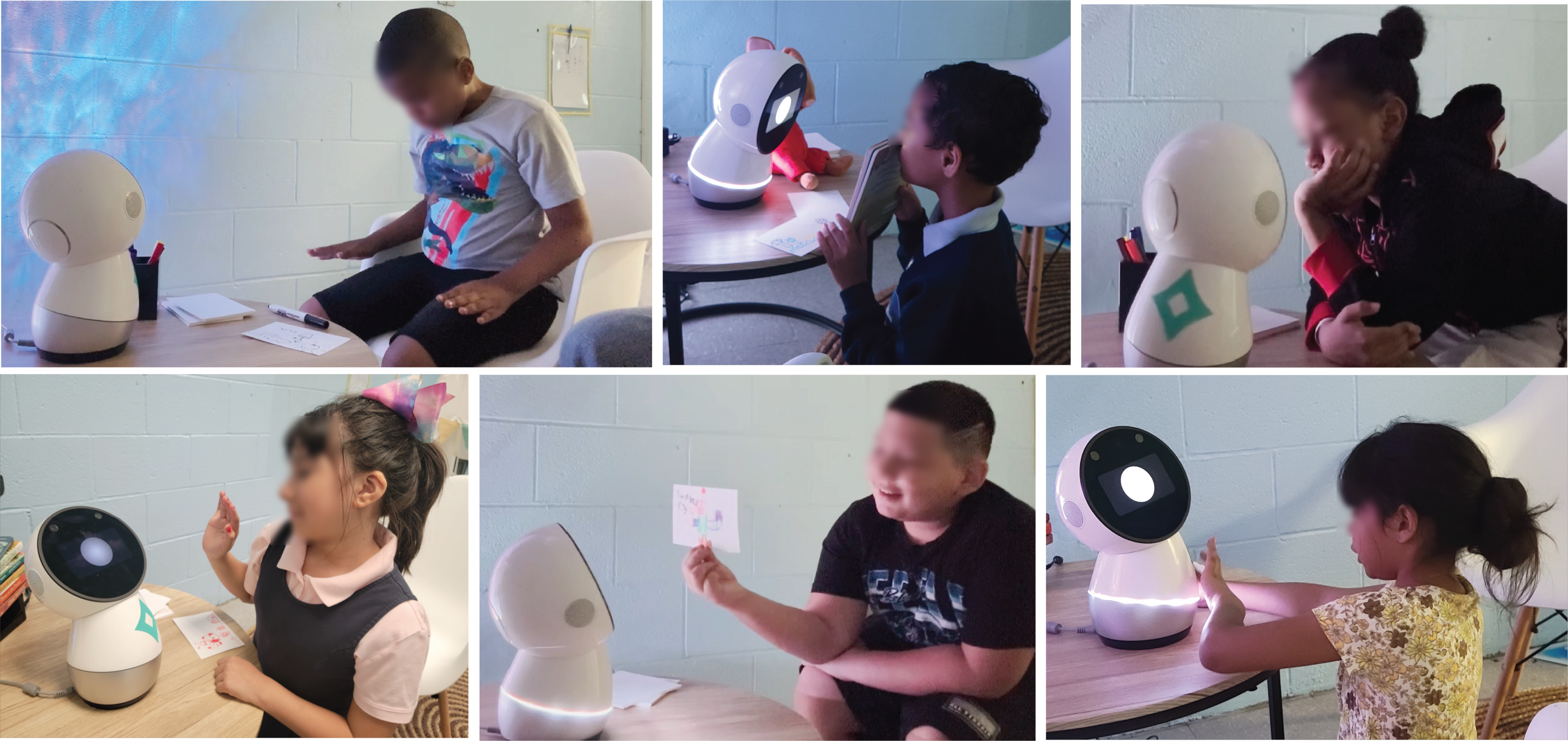}
    \caption{\textbf{Observer-Mediated Robot-Assisted Activities in a School-Based De-Escalation Setting.} The robot engages students in structured routines (e.g., small talk, guided breathing, and collaborative tasks), each governed by an activity-specific observer and overlay set. A higher-level supervisory observer regulates when to initiate, terminate, or switch activities based on the student's interaction state, enabling state-dependent control over both content and timing. Implementation details are further described in \citeay{ramnauth2025reset}.}
    \label{fig:reset}
\end{figure}

\subsubsection{Observer Configuration and System Architecture}
RESET implemented a hierarchical Grounded Observer architecture that supervised both utterance-level generation and activity-level control. At the lowest level, each intervention activity was paired with an activity-specific observer and overlay set that enforced constraints appropriate to that mode (e.g., pacing and affect for deep-breathing guidance, or open-ended encouragement but not instructive feedback during collaborative tasks). These observers regulated the robot's utterances and timing within each activity, ensuring that the robot remained within the bounds of the intended interaction structure even when the student's behavior was unpredictable. Because each activity had a distinct structure and success criterion, each overlay set encoded a different admissible region of behavior.


Above these activity-level observers, RESET incorporated a higher-level oversight observer that governed \textit{mode switching}. This higher-level supervisor evaluated the interaction state to determine whether the robot should initiate engagement, remain silent, terminate an activity, or transition to a different routine entirely. For instance, it could suppress small talk when verbal engagement was unproductive, shift from conversation to breathing when fixation was detected, or stop prompting when the student repeatedly refused to respond. In the formalism introduced earlier, this oversight observer effectively constrained the admissible action set at the level of high-level policies, selecting among different sub-policies whose admissibility conditions differed. This hierarchical structure mirrors robotics architectures in which a supervisory controller selects among task controllers based on state, while each controller enforces its own safety constraints locally.

A key guardrail design choice in RESET was the separation between \textit{content generation} and \textit{interaction regulation}. Because activity content was templated, the primary locus of runtime control shifted toward regulating the \textit{timing} of the interaction content. The observer therefore operated heavily on temporal and contextual features. The most salient features for the system's decision-making were not lexical properties of text, but interaction-level signals such as prolonged silence, repeated refusal, rapid topic cycling, fixation, and indicators that escalation was worsening rather than improving.

The deployed system also introduced practical infrastructure constraints. School deployments required reliability under frequent interruptions, constrained network access, and minimal burden on staff. The architecture therefore emphasized robustness and recoverability, such as autonomous startup routines and conservative defaults when state could not be confidently inferred.

\subsubsection{Lessons for Guardrails Under State Volatility}
RESET highlights a class of the larger guardrail problem that is not well captured by standard notions of ``safe text generation.'' The primary challenge is not preventing isolated unsafe outputs, but maintaining stable behavior when the interaction state is volatile, partially observed, and capable of shifting rapidly. In this context, the admissible action set is not only state-dependent but effectively non-stationary: as a student transitions from mild frustration to acute dysregulation, the set of socially acceptable actions can collapse abruptly. Behaviors that are appropriate when the student is calm (e.g., open-ended questions, playful elaboration) can become destabilizing when the student is overwhelmed. This implies that guardrails for crisis-adjacent interaction must be explicitly phase-conditioned, and that constraint enforcement must be sensitive not only to the current state estimate but to the possibility of rapid state shifts. From a dynamical systems perspective, RESET illustrates that safety cannot be treated as a static property of responses; it is a property of the closed-loop trajectory, and the safe set itself may shrink as the system approaches certain states.


A second lesson is that conventional alignment objectives such as ``helpfulness'' become unreliable in these contexts. In high-arousal states, an agent's default tendency to be maximally informative and always responsive can increase cognitive load and prolong dysregulation. The stabilizing action is often not to provide more content, but to reduce stimulation such as through shorter utterances, fewer questions, slower pacing, or deliberate silence. In RESET, many interventions were deliberately designed to be low-demand (and ignored, if needed), and overlays were tuned to suppress behaviors that are typically rewarded in language model optimization (response accuracy and specificity, proactive advice). This reveals an important misalignment between the objectives that foundation models are trained to optimize and the objectives required for safe intervention. In crisis-adjacent settings such as this, guardrails must therefore enforce constraints not only on content safety, but on interaction \textit{demand level}---treating cognitive burden and conversational pressure also as quantities that must be aligned.

Third, RESET demonstrates that safety in socially sensitive domains sometimes requires controlled violations of what is expected of the system itself. In most dialogue systems, interrupting a user mid-sentence or refusing to continue a line of conversation is treated as undesirable. In this context, however, these behaviors were necessary for a successful intervention \cite{ramnauth2026rules}. The robot occasionally needed to interrupt perseverative loops, terminate unproductive conversation, or force a transition into a different activity. This highlights a broader implication for guardrail design: constraint satisfaction cannot always be equated compliance. Instead, the safe set must include a notion of \textit{protective override}, in which the system is permitted---or even required---to violate surface-level social conventions in order to prevent trajectories that are likely to lead to harm. This is directly analogous to supervisory control in robotics, where a safety controller overrides a nominal controller when the system approaches unsafe regions.


RESET also illustrates that in time-sensitive social interventions, the safe set is shaped by real-time feasibility constraints. Latency is not merely a usability concern; it is a behavioral variable that affects interpretation and escalation. A response that is semantically appropriate but delayed can be perceived as dismissive or socially incoherent, and can therefore worsen the interaction state. This suggests that guardrail architectures for embodied, real-world deployments must explicitly incorporate temporal constraints into admissibility. More broadly, it implies that runtime enforcement cannot be evaluated solely in terms of correctness or safety of generated content; it must be evaluated as a control problem under computational and temporal budgets.

A further lesson is that hierarchical supervision is not simply an engineering convenience but an enabling safety mechanism. RESET's separation between activity selection and activity execution provided a structured way to constrain behavior while still adapting to unpredictable student responses. At the observer level, this decomposition yielded two distinct enforcement problems: selecting which sub-policy is admissible at the current state, and ensuring that behavior within that sub-policy remains bounded. This structure provides a clearer mapping between specification and execution, and it enables fault localization when failures occur. If an interaction becomes unproductive, the failure can be attributed either to incorrect mode selection (an oversight-level violation) or to incorrect execution within a mode (an activity-level violation). This kind of diagnosability is central to safety engineering in robotics, and RESET demonstrates its value for foundation-model-driven systems operating in complex social environments.

Finally, the deployment emphasizes that in institutional settings the safe set is defined not only by the primary user, but also by the surrounding human stakeholders. Teachers and staff were not passive observers but part of the broader closed-loop system, and the robot needed to remain appropriate and accountable within school routines. This suggests that robust guardrails must incorporate requirements of governance and operational control, not merely content safety. RESET thus highlights that deploying foundation models in real-world institutions demands an expanded notion of constraint specification: one that includes not only what the model may say, but when it may engage, how it may intervene, and how its behavior can be audited and overridden by human stakeholders. 


\section{Discussion}
This paper has argued that efforts to build guardrails for foundation models can be usefully reframed through a robotics lens. Rather than treating safety and alignment as properties that must be absorbed into a model through training alone, we emphasized a control-oriented view in which deployed foundation model systems are dynamical systems operating over interaction trajectories. In this framing, safety is not reducible to whether a model produces a locally acceptable output, but instead concerns whether the system remains within admissible regions of behavior over time under uncertainty, partial observability, and shifting human demands through interaction. The primary contribution of the paper is therefore conceptual: we introduced a constrained dynamical systems formalism for socially interactive foundation model systems, identified concrete technical constructs from robotics that operationalize this view, and presented the Grounded Observer framework as a practical instantiation that preserves an explicit locus of runtime control.

\subsection{Why Current Guardrail Paradigms Are Incomplete}
A key implication of this framing is that many widely deployed guardrail mechanisms for foundation models are incomplete when applied to socially sensitive domains. Training-time alignment methods such as RLHF and instruction tuning can shape a model's distribution of behaviors, but do not provide a mechanism for enforcing constraints at runtime once the model is deployed. Similarly, prompt engineering and system messages can encode high-level behavioral instructions, but cannot reliably prevent gradual drift across long-horizon interactions. Post hoc content filtering, while useful for removing overtly unsafe outputs, remains fundamentally reactive and often operates without access to the latent conversational state that determines whether an action is appropriate. These approaches are often evaluated using single-turn benchmarks that assume safety can be assessed locally. Yet many of the most consequential failures in socially interactive systems are not catastrophic in a single step; they accumulate. Drift toward an inappropriate conversational motive, escalation of affect, subtle shifts in role framing, or repeated small violations of user-specific norms may remain individually plausible while jointly producing an interaction trajectory that is socially misaligned. In this sense, the technical lesson from robotics is that constraint satisfaction must be understood as an ongoing property of closed-loop system evolution.

The constrained dynamical systems view provides a vocabulary for making these issues explicit. By separating state, action, and admissibility, the framework clarifies what is often left implicit in discussions of foundation-model-safety: the safety of an output cannot be evaluated independently of the state representation used to contextualize it, and guarantees cannot be stronger than the fidelity of the features and observers that define the safe set. This is not a weakness of the formalism but an important point of transparency. In robotics, safety claims are always conditional on sensing and state estimation. Similarly, in foundation model systems, any behavioral guarantee is conditional on what the system can reliably measure about the interaction, the user, and the environment. The value of the formalism is that it forces these dependencies to be explicitly audited, rather than hidden inside broad notions of ``alignment'' or ``helpfulness.''

\subsection{Scope of the Grounded Observer Framework and Toward Stronger Guarantees}
The Grounded Observer framework was presented as one instantiation of this formal structure. The framework is not intended as a new safety primitive, but as an example of how robotics-inspired runtime enforcement can be realized for foundation model systems. By explicitly decomposing a deployed system into a base policy and an external observer, the framework provides a locus of control that is absent from purely training-based approaches. Constraints are represented externally as overlays that can be modified, composed, or revoked without retraining the base model, and the observer enforces admissibility at runtime by filtering candidate actions. Importantly, the framework supports constraint enforcement over trajectories rather than single-turn responses, allowing the system to respond to longitudinal properties such as repeated violations or shifting interaction goals. The framework therefore makes a concrete argument: even when the underlying model remains stochastic and unconstrained, the deployed system can be structured such that constraint enforcement is an explicit part of the action-selection loop.

At the same time, the Grounded Observer framework has clear limitations that constrain the strength of any behavioral guarantees it can provide. Most importantly, the observer enforces constraints only to the extent that the relevant properties of interaction can be captured by feature extractors and state representations. If the system cannot reliably detect user affect, social context, or subtle pragmatic intent, then constraint enforcement may fail silently. Similarly, the overlays themselves may be incomplete or mis-specified. This problem is not unique to our framework; it is inherent to any attempt to operationalize social norms as evaluable constraints. However, the framework makes this limitation explicit by isolating where the uncertainty resides: not in the base model alone, but in the sensing and specification layers that define admissibility. This suggests that future progress in guardrails will depend not only on better language models, but on better interaction state estimation and more principled methods for authoring, validating, and adapting constraint specifications.

\subsection{Interpreting the Application Studies and Their Limitations}
The three application studies---small talk, in-home autism therapy, and school de-escalation---were included not as definitive validation of the observer architecture, but as demonstrations of how the robotics constructs outlined in Section~\ref{sec:constructs} manifest in real-world, socially sensitive deployments. A natural critique is that these case studies do not prove that the observer yields optimal performance, or that it outperforms alternative alignment techniques under controlled benchmarking. This is correct. The purpose of these applications is instead to illustrate that socially sensitive domains often require guardrails that operate on trajectory-level stability, that regulate not only what is said but when engagement is appropriate, and that impose different admissibility criteria depending on interaction phase. These demands are difficult to satisfy using static prompting or training-time alignment alone, because the relevant constraints are state-dependent, compositional, and must be enforced repeatedly under real-world uncertainty. The case studies therefore serve as existence proofs: they demonstrate that the formal decomposition into state, admissible actions, and runtime enforcement is not merely theoretical, but can be instantiated in autonomous systems interacting with real users over extended periods.

These deployments also motivate a broader argument about what constitutes ``success'' in guardrail design for socially interactive AI. Unlike functional tasks with objective performance metrics, many socially embedded goals are evaluated subjectively, relationally, and retrospectively. In small talk, the failure mode is rarely a single unacceptable utterance, but the gradual loss of conversational coherence and motive. In in-home therapy, the primary concern is not fluency but psychological safety and consistency of pedagogical framing. In de-escalation, the same conversational move may be appropriate in one phase and destabilizing in another, requiring state-dependent constraint enforcement and careful regulation of interaction demand. These domains are therefore representative of a class of problems that are increasingly central to real-world deployment of foundation models, yet are poorly captured by current benchmark paradigms. The relevance of these applications is not that they represent all guardrail settings, but that they highlight important and nuanced subclasses of guardrail problems.

One may raise the question of whether our model choices in our deployment studies are presently relevant. The specific deployments relied on particular foundation models available at the time (e.g., models prior to GPT 4o), and it is reasonable to ask whether newer or more capable models would reduce the need for such supervisory architectures. While improved base models may reduce certain failure modes, the broader control problem remains. As models become more capable and open-ended, their capabilities expand the action space and thereby increase the surface area over which behavioral constraints must be enforced. In this sense, the need for runtime guardrails is not eliminated by stronger models; it is amplified. The observer framework is therefore intended to speak generally across model classes: it is a system-level architectural approach that remains applicable even as the base policy improves.

Lastly, the deployments underscore that guardrails in socially sensitive domains cannot be fully decoupled from embodiment and real-world infrastructure. In embodied systems, timing, pacing, gaze, and physical presence alter the interpretation of language, and failures are attributed to the agent as a social entity rather than to the underlying model as software. In long-term home and school deployments, safety includes operational reliability: failures in sensing, connectivity, or scheduling can indirectly produce unsafe interaction patterns even when the language model itself is well-behaved. This suggests that guardrail research should expand beyond model-centric mechanisms and consider system-level constraints that preserve stable closed-loop operation under real-world uncertainty. 

Such concerns are underemphasized in current research, which often focuses on output content safety while assuming that interaction itself is always desirable. In contrast, robotics and human-centered autonomy have long emphasized that safe behavior includes deciding when \textit{not} to act.

\section{Conclusion}
In all, this paper argues that robotics offers more than inspiration or analogy for foundation model safety. It provides mature technical constructs---runtime supervision, explicit constraint specification, state-dependent admissibility, hierarchical control, and closed-loop reasoning---that can be imported to strengthen guardrails for socially interactive foundation model systems. The Grounded Observer framework demonstrates one way of instantiating these constructs in practice while preserving a generalizable formal structure. Although the framework does not provide absolute guarantees, it offers a concrete step toward moving safety beyond distributional alignment and toward enforceable behavioral bounds. As foundation models are increasingly deployed in settings where interaction unfolds over time and failures accumulate gradually, such runtime control architectures may become a necessary complement to existing approaches.


\printbibliography
\end{document}